\pdfoutput=1

\documentclass{article} %
\usepackage{iclr2027_conference,times}

\usepackage{amsmath,amsfonts,bm}

\def\eqref#1{equation~\ref{#1}}

\def\1{\bm{1}}

\DeclareMathAlphabet{\mathsfit}{\encodingdefault}{\sfdefault}{m}{sl}
\SetMathAlphabet{\mathsfit}{bold}{\encodingdefault}{\sfdefault}{bx}{n}

\usepackage{amsmath, amssymb, MnSymbol}
\usepackage{amsthm}
\newtheorem{proposition}{Prop.}
\theoremstyle{definition}
\newtheorem*{problemstatement}{Problem statement}
\newtheorem{limitation}{Limitation}
\newtheorem{assumption}{Assumption}
\theoremstyle{plain}
\usepackage{hyperref}
\usepackage{url}
\usepackage{xcolor}
\usepackage{numprint}

\definecolor{primary}{HTML}{1B5E43}      %
\definecolor{secondary}{HTML}{1F4E79}    %
\definecolor{tertiary}{HTML}{5B3A7E}     %
\definecolor{quaternary}{HTML}{8A5A1B}   %
\definecolor{quinary}{HTML}{8C2F39}      %
\definecolor{senary}{HTML}{146B6B}       %
\definecolor{primarytext}{HTML}{FFFFFF}  %
\definecolor{highlight}{HTML}{FAE49C}    %
\definecolor{muted}{HTML}{8A8A8A}        %
\definecolor{border}{HTML}{B0B0B0}       %
\usepackage{csquotes}
\usepackage{booktabs}
\usepackage{graphicx}
\usepackage{float}
\usepackage{tikz}
\usepackage{threeparttable}
\usepackage{adjustbox}  %
\usepackage{caption}  %
\usepackage{etoc}  %
\usepackage{subcaption}
\usetikzlibrary{arrows.meta,positioning,fit}

\def\rowstep{2.9}   %
\def\namegap{1.7}   %

\newlength{\cellw}
\newlength{\cellh}
\newcommand{\relcell}[5]{%
	\node[#1, anchor=west, minimum width=\csname relw@#3\endcsname]
	(#2) at (#4) {#5};%
}

\newcommand{\relcells}[3]{%
	\gdef\relanchor{#2}%
	\foreach \lbl/\sty [count=\i from 0] in {#3}
		{\expandafter\relcell\expandafter{\sty}{#1\i}{\i}{\relanchor}{\lbl}%
			\xdef\relanchor{#1\i.east}}%
}

\newcommand{\relwidths}[1]{%
	\foreach \w [count=\j from 0] in {#1}
		{\expandafter\xdef\csname relw@\j\endcsname{\w mm}}%
}

\newcommand{\reltable}[3]{%
	\def\tabx{#1}\def\taby{#2}%
	\foreach \j in {0,...,9}
		{\expandafter\xdef\csname relw@\j\endcsname{\the\cellw}}%
	\node[tname] at (\tabx, \taby+\namegap) {#3};%
}

\newcommand{\relrow}[3]{\relcells{#1}{\tabx,\taby-#2*\rowstep}{#3}}

\newcommand{\panellabel}[5]{%
	\node[fit=#3, inner sep=0pt, draw=none] (#2) {};%
	\node[panel] at (#2.center |- 0,#4)
		{\phantomsubcaption\label{#1}(\thesubfigure) #5};%
}

\colorlet{refcolor}{green!55!black}
\hypersetup{
  colorlinks=true,
  linkcolor=red,        %
  citecolor=refcolor,   %
  urlcolor=refcolor,    %
}

\usepackage[nameinlink]{cleveref}
\crefname{proposition}{Prop.}{Props.}
\Crefname{proposition}{Prop.}{Props.}
\crefname{limitation}{Lim.}{Lims.}
\Crefname{limitation}{Lim.}{Lims.}
\crefname{assumption}{Asm.}{Asms.}
\Crefname{assumption}{Asm.}{Asms.}
\crefname{figure}{Fig.}{Figs.}
\Crefname{figure}{Fig.}{Figs.}
\crefname{subfigure}{Fig.}{Figs.}
\Crefname{subfigure}{Fig.}{Figs.}
\crefname{table}{Tab.}{Tabs.}
\Crefname{table}{Tab.}{Tabs.}
\crefname{section}{Sec.}{Secs.}
\Crefname{section}{Sec.}{Secs.}
\crefname{subsection}{Sec.}{Secs.}
\Crefname{subsection}{Sec.}{Secs.}
\crefname{appendix}{App.}{Apps.}
\Crefname{appendix}{App.}{Apps.}

\title{%
  RelICL: Training-free Relational Learning with Tabular Foundation Models
}

\author{Simon Forbat \& Rainer Gemulla \\
 Data and Web Science Group\\
 University of Mannheim\\
 Mannheim, Germany \\
  \texttt{\{sforbat,rgemulla\}@uni-mannheim.de}
}

\newcommand\sset[1]{\{#1\}}

\newcommand{\tfm}{\mathcal{T}}
\newcommand{\tenc}{\mathcal{T}_{\mathrm{enc}}}
\newcommand{\tpred}{\mathcal{T}_{\mathrm{pred}}}

\newcounter{contribution}
\renewcommand{\thecontribution}{C\arabic{contribution}}
\newcommand{\contribution}{%
	\refstepcounter{contribution}\textbf{(\thecontribution)}}

\newif\ifpreprint
\preprinttrue
\iclrfinalcopy
\begin{document}

\maketitle
\lhead{Preprint.}
\etocdepthtag.toc{main}

\begin{abstract}
  Tabular foundation models achieve state-of-the-art performance on single-table
  tasks without any training. Recent work suggests that they are also
  well-suited for relational learning via deep feature synthesis (DFS), which
  flattens a relational schema into a single table by adding aggregates of the
  other tables' columns as features. This approach is appealing because it
  directly benefits from improvements to or customization of the underlying
  tabular foundation model. In this paper, we identify two key problems with
  DFS: \emph{feature explosion} and \emph{interaction blindness}. The first
  problem arises because the number of DFS features grows quickly as the schema
  becomes more complex, limiting scalability and performance. The second problem
  arises because column-wise aggregates do not account for feature interactions,
  limiting performance. We propose and explore an alternative method termed
  RelICL, which keeps the benefits of DFS but alleviates these two problems. At
  its heart, RelICL propagates and fuses information step by step through the
  schema graph, using the same tabular foundation model that is eventually used
  for prediction to do so. In our experimental study using RelBench tasks,
  RelICL was on par with the strongest approach based on deep feature synthesis.
\end{abstract}

\section{Introduction}

Relational data is ubiquitous in industry and science \citep{Fey2024}, spanning
application areas like e-commerce, social networks, and clinical research
\citep{Robinson2024}. A relational database can be understood as a collection of
multiple tables, interconnected with each other via foreign key references
\citep{Codd1970,Garcia-Molina2009}. In this paper, we explore how to leverage
strong tabular foundation models for relational learning.

Tabular foundation models (TFMs) such as TabPFN \citep{Hollmann2022,
	Hollmann2025a}, TabICL \citep{Qu2025, Qu2026}, and others \citep{Ma2025,
	Zhang2025, Spinaci2025, Kong2026} are pretrained on a large
amount of synthetic and/or real-world tabular data. They reach state-of-the-art
performance for single-table learning tasks using in-context learning, i.e.,
without further training. For example, on the tabular benchmark TabArena
\citep{Erickson2025}, most top-ranked models\footnote{Latest leaderboard:
	\url{https://huggingface.co/spaces/TabArena/leaderboard}} are TFMs,
outperforming gradient-boosted trees such as XGBoost \citep{Chen2016}, deep
tabular models such as RealMLP \citep{Holzmuller2024}, and AutoML systems such
as AutoGluon \citep{Erickson2020a}.

A natural and practically important question is whether and how TFMs can be
leveraged for relational learning on a set of interconnected tables. More
specifically, given a \emph{task table} holding in-context examples and
prediction targets, how can we leverage information from the remaining
\emph{context tables}? \emph{Deep feature synthesis (DFS)} \citep{Kanter2015}
provides one answer to that question: It extends the task table with additional
\emph{DFS features}, each obtained by aggregating values from a single context
table column, and then applies a TFM to the resulting \emph{extended task
	table}. To do so, DFS follows \emph{meta-paths} \citep{Sun2011}, i.e., chains
of foreign key references that start at the task table and end at a context
table. For every such meta-path, DFS aggregates the context table's columns
using multiple aggregation functions (e.g., mean, sum, median, mode, minimum,
maximum). For example, in a social network task, a DFS feature may hold the
average age of the friends of a given user. DFS needs no training, it can be
used with any tabular model, including TFMs, and it can reach state-of-the-art
results for relation learning~\citep{Xu2026a, Hayler2026}. DFS is implemented in
several frameworks such as Featuretools \citep{Kanter2015}, RDBLearn
\citep{Xu2026a, Zhang2026} and TabPFN-Rel \citep{Hayler2026}.

Despite its simplicity and strong empirical performance, applying DFS in practice
can be challenging due to \emph{feature explosion}: the number of DFS features
grows quickly as the schema becomes more and more complex. Since DFS aggregates
every column along every meta-path using multiple aggregation functions, the
width of the extended task table grows with the number and width of the context
tables, limiting scalability. For this reason, practical DFS approaches need to
use strategies such as capping meta-paths length or pruning DFS
features~\citep{Kanter2015, Lam2017, Xu2026a, Hayler2026}, limiting performance.
A second inherent limitation of DFS is \emph{interaction blindness}: since each
DFS feature is an aggregate of a single column, DFS cannot capture interactions
between the columns the context tables. Any task that relies on such
interactions is therefore impossible to solve with DFS.

In this paper, we explore the question of whether we can keep the benefits of
DFS, i.e., training-free relational prediction using any tabular foundation
model, while avoiding its drawbacks of feature explosion and interaction
blindness. To do so, we propose a simple approach termed RelICL. Like DFS,
RelICL builds an extended task table fed into a TFM for the final predictions.
Unlike DFS, RelICL also exploits the TFM to build the extended task table. In
more detail, the TFM is additionally used for \emph{feature extraction} (row
embeddings from the context tables) and \emph{feature fusion} (propagating row
embeddings step by step). RelICL does not suffer from feature explosion because
the number of additional features only depends on the size of the local
neighborhood instead of on the full database. RelICL does not suffer from
interaction blindness because it propagates, pools, and fuses row embeddings
instead of individual context-table columns. RelICL works out of the box with
any TFM which provides row embeddings, e.g., most recent TFMs including TabICL
\citep{Qu2026}, TabPFN \citep{Hollmann2025a}, or TabFM~\citep{Kong2026}.

To summarize, the contributions of this paper are: \contribution{} We identify
two major limitations of DFS, feature explosion and interaction blindness
(\cref{sec:dfs}) and show that there are tasks for which DFS-based predictors
can do no better than random guessing. \contribution{} We propose RelICL
(\cref{sec:relicl}), which avoids both limitations by replacing column-wise
aggregation along meta-paths with propagation and fusion of row embeddings along
individual foreign key references. \contribution{} We report on an experimental
study (\cref{sec:experiments}) using the 21 entity tasks of RelBench
\citep{Robinson2024, Gu2026a}, indicating that RelICL is on par with the
strongest DFS-based approach (and outperforming many alternative approaches) and
that the underlying TFM can be easily swapped.
Source code and results are available at
\url{https://github.com/uma-pi1/relicl}.

\section{Deep Feature Synthesis and its Limitations}
\label{sec:dfs}

\subsection{Preliminaries}

A relational database $D$ is a set of tables interconnected by references, see
the example in \cref{fig:method-input}. We follow \citet{Fey2024, Robinson2024} and
model relational learning via a \emph{task table} $T$, which holds in-context
examples ($T_\text{train}$, green in \cref{fig:method-input}) and \emph{query rows}
($T_\text{eval}$, yellow), and multiple \emph{context tables}, which hold
auxiliary information. A \emph{reference} links the rows of two tables via key
columns: if two rows have the same key value, they are linked together. For
example, in \cref{fig:method-input}, there is a reference between \emph{lap times}
(key \emph{r\_id}) and \emph{results} (key \emph{r\_id}). References can be
one-to-one, many-to-one (e.g., \emph{lap times} and \emph{results}) or
many-to-many (e.g., \emph{task} and \emph{results}). Tables and references can
be represented via a \emph{schema graph}.
\begin{problemstatement} Given $D$, $T_\text{train}$, and $T_\text{eval}$, use a
	pretrained tabular foundation model $\tfm$ to predict the target attribute of
	the rows in $T_\text{eval}$.
\end{problemstatement}

A related task is relational deep learning~\citep{Fey2024, Robinson2024}, which
adds a temporal component for tasks such as forecasting. Here predictions at
time $t$ must not be influenced by data newer than $t$. For example, to predict
whether Cathi wins the next race in the example of \cref{fig:method-input}, we
should not already be aware of the results of the next race. We
follow~\citet{Fey2024, Robinson2024} and address such tasks by temporal
filtering using a cut-off time (gray ``future'' rows in \cref{fig:method-input}) and
temporal pooling (discussed later). In most of the following discussion, we
ignore the temporal component for ease of exposition.

Our problem statement specifically requires the use of a single TFM.
Approaches such as graph neural
networks~\citep{Fey2024, Robinson2024,Chen2025,Dwivedi2025b},
which are trained per database, and relational foundation
models~\citep{Ranjan2025a, Kothapalli2026}, which are pretrained on a relational
database corpus, fall outside this setting. We focus on TFMs because they
achieve state-of-the-art single-table performance, large corpora of tabular data
are readily available~\citep{Gardner2024}, TFMs are actively developed and
continuously improved, and TFMs are easy to customize for a particular domain
via continual pre-training. Any advances in TFMs thus directly translate to
advances in relational learning based on TFMs.

\subsection{Deep Feature Synthesis}

Deep feature synthesis \citep{Kanter2015} \emph{extends} the task table $T$ by
DFS features, i.e., columns derived from the context tables along meta-paths,
and uses the resulting extended task table $T_\text{DFS}$ for prediction. A
\emph{meta-path} $P=(T, R_1, \dots, R_L)$ of length $L$ is a sequence of $L+1$
tables starting at the task table $T$ such that every two consecutive tables are
connected by a reference. For example, (\emph{task}, \emph{results}, \emph{lap
	times}) is a length-2 meta-path in \cref{fig:method-input}. Given a meta-path,
DFS determines for every row $x$ in $T$ the set of rows $P(x)$ in $R_L$ that are
reachable from $r$ by following the meta-path $P$.\footnote{For relational deep
	learning, only rows that are reachable without accessing future rows are
	considered.} The columns in $P(x)$ are then aggregated individually using multiple
aggregation functions, e.g., length, mean/min/max/std (for numerical columns),
mode/value frequencies (for categorical columns). The resulting aggregates then
form the DFS features for meta-path $P$. This process is repeated for each
meta-path. The number of meta paths is at least linear but can be exponential
with respect to the number of tables in the database \citep{Lam2017}.

Once the extended task table has been obtained, any tabular model can be used
for prediction. This includes classical models such as XGBoost \citep{Chen2016},
CatBoost \citep{Prokhorenkova2018}, or LightGBM \citep{Ke2017}, deep tabular
models such as RealMLP \citep{Holzmuller2024} or FT-Transformer
\citep{Gorishniy2021}, and tabular foundation models as in the recent approaches
RDBLearn \citep{Xu2026a, Zhang2026} and TabPFN-Rel \citep{Hayler2026}.

\subsection{Feature Explosion}

\begin{limitation}[Feature explosion]
	\label{lim:feature-explosion}
	The number of DFS features grows with the complexity of the schema.
\end{limitation}

The extended task table holds one column per combination of (i) meta-path, (ii)
column of the context table where that meta-path ends, and (iii) aggregation
function. Its width is therefore $\mathcal O(\#\text{meta-paths} \cdot
	\#\text{attributes} \cdot \#\text{aggregates})$. For instance, the RelBench database
rel-f1 \citep{Robinson2024} holds nine tables with 67 columns in total
(\cref{app:datasets}). RDBLearn generates \numprint{7715} columns on the
\texttt{driver-top3} task with its six default aggregation functions (before
pruning down to \numprint{5460}, unbounded depth), more than one hundred times
the width of the schema itself.

In practice, this problem needs to be addressed. Some works do this by selecting
a small set of meta-paths manually \citep{Wang2019}, whereas others aim to learn
promising meta-paths \citep{Ferrini2024}. The perhaps most common approach,
however, is to bounds the maximum length of the considered meta-path, the
\emph{depth} \citep{Kanter2015, Xu2026a, Hayler2026}. In fact, the latest
versions of featuretools (1.31.0) \citep{Kanter2015}, RDBLearn (1.1)
\citep{Xu2026a}, and TabPFN-Rel \citep{Hayler2026} all use a maximum depth of 2
by default. Such an approach can be highly problematic, however.

\begin{proposition}[Insufficiency of bounded depth]
	\label{prop:depth}
	For every depth bound $D$, there are relational prediction tasks that no
	DFS-based approach can solve better than random guessing.
\end{proposition}

This problem arises when the relevant signal for prediction is too far away from
the task table, and it holds no matter how powerful the tabular model or the
considered aggregation functions are. As a simple example, consider a task where
the target label is directly available as a feature in a context table reachable
from the task table by following a meta-path of length 3. Any DFS-based approach
with depth bound 2 cannot find this signal and thus fails. \cref{tab:dfs-depth}
shows empirical results for this task (1000 training rows, 300 query rows, 3
seeds) using RDBLearn (v1.1, default configuration, i.e., depth bound 2) and
RelICL (\cref{sec:relicl}) with both TabICL and TabPFN as TFM backbones. RDBLearn
reached a test AUROC of 50.8\% on average, i.e., random guessing. RelICL
consistently reached 100.0\% AUROC with both backbones. See \cref{app:depth} for
details.

In addition to bounding meta-path length, prior work explored post-hoc feature
selection methods, e.g., through information gain
\citep{Markovitch2002,Piramuthu2009,Fan2010}, through feature importance
\citep{Shi2020}, or through tests of independence with respect to the target
\citep{Lam2017}. These methods milden the problem, but they also require
computation of all DFS features first and they ultimately may remove relevant
information from the extended task table.

\begin{table}[t]
	\centering
	\small
	\begin{minipage}[t]{0.48\linewidth}
		\centering
		\caption{%
			Bounded depth task of \cref{prop:depth}, with \mbox{RDBLearn} at its default
			depth bound of two (AUROC (\%), higher is better, mean $\pm$ SEM over
			three seeds).
		}
		\label{tab:dfs-depth}
		\begin{tabular}{lr}
\toprule
\textbf{Method} & \textbf{AUROC} \\
\midrule
RDBLearn (depth bound 2) & 50.8 $\pm$ 0.1 \\
RelICL (TabICL, unbounded) & 100.0 $\pm$ 0.0 \\
RelICL (TabPFN, unbounded) & 100.0 $\pm$ 0.0 \\
\bottomrule
\end{tabular}

	\end{minipage}\hfill
	\begin{minipage}[t]{0.48\linewidth}
		\centering
		\caption{%
		Interaction blindness task of \cref{prop:single-column} (AUROC (\%),
		higher is better, mean $\pm$ SEM over three seeds).
		}
		\label{tab:dfs-interaction}
		\begin{tabular}{lr}
\toprule
\textbf{Method} & \textbf{AUROC} \\
\midrule
RDBLearn & 46.5 $\pm$ 3.3 \\
RelICL (TabICL) & 99.3 $\pm$ 0.7 \\
RelICL (TabPFN) & 99.8 $\pm$ 0.1 \\
\bottomrule
\end{tabular}

	\end{minipage}
\end{table}

\subsection{Interaction Blindness}

\begin{limitation}[Interaction blindness]
	\label{lim:interaction-blindness}
	DFS-based methods are oblivious to feature interactions.
\end{limitation}

Since each DFS feature is an aggregate of a single context-table column,
DFS-based approaches are inherently oblivious to feature interactions. In fact,
\citet{Xu2026a} makes single-column aggregation their \enquote{guiding
	principle} according to which aggregates should compress only within but not
across columns. Again, such an approach can be highly problematic.

\begin{proposition}[Insufficiency of DFS features]
	\label{prop:single-column}
	There are relational prediction tasks that no DFS-based predictor can solve
	better than random guessing.
\end{proposition}

This arises when the signal required to solve the task lies solely in feature
interactions, but not feature values. Consider, for example, a relational
learning task with one context table, in which every task table row references
exactly three rows in the context table The three rows are either
$\{(0,-1),(1,0),(2,1)\}$ (positive correlation) or $\{(0,1),(1,0),(2,-1)\}$
(negative correlation), and the task is to predict which of the two cases hold.
Since each column has the same set of values, all DFS features necessarily
coincide for both cases so that a DFS-based method cannot distinguish between
them. \cref{tab:dfs-interaction} validates this empirically (1000 training rows,
300 query rows, 3 seeds): RDBLearn (v1.1, default configuration) is no better
than random guessing, whereas RelICL (\cref{sec:relicl}) has good performance.
\Cref{app:cross-column} provides further details.

One way to combat interaction blindness is to engineer additional features,
e.g., manually or via an LLM (\cref{app:related-work}). Such an approach is
challenging to get right, however, and may lead to further feature explosion.

\section{RelICL}
\label{sec:relicl}

\begin{figure}[t]
	\centering
	\input{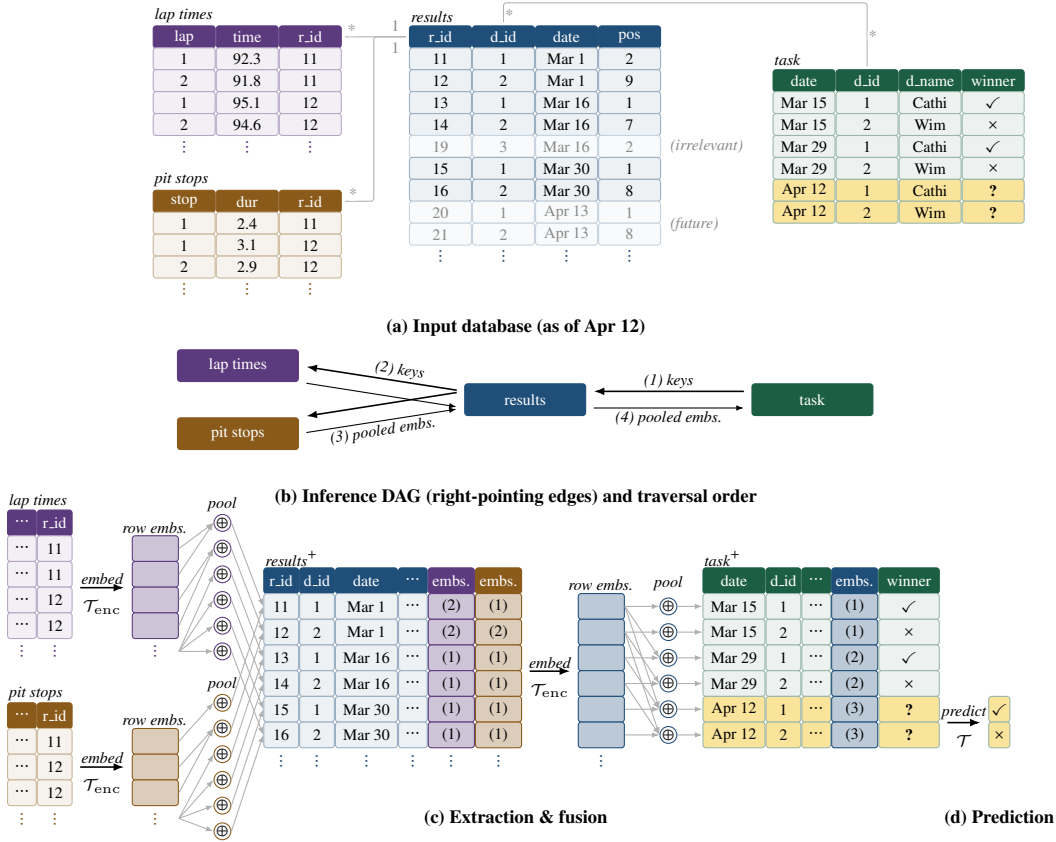}
	\caption{%
		A binary classification task loosely inspired by \texttt{driver-top3}
		\citep{Robinson2024}, predicting the drivers who win the next race.
		Highlighted rows are the ones being predicted.
		\textbf{\subref*{fig:method-input}} The input database.
		Grayed-out rows are removed before any modeling because (i) they either fall
		after the evaluation cutoff or (ii) they are irrelevant for the task.
		\textbf{\subref*{fig:method-dag}} The inference DAG
		(\cref{sec:extraction}), with tables ordered by their distance from the
		task table.
		The outward traversal (upper arrows) carries the keys that mark the
		relevant rows.
		The inward traversal (lower arrows) in turn carries the pooled row
		embeddings.
		\textbf{\subref*{fig:method-fusion}} Each table pools the row embeddings of its children and gains one further column block per child.
		\textbf{\subref*{fig:method-pred}} Predict on the task table extended
		with the pooled row embeddings from \textbf{\subref*{fig:method-fusion}}.
	}
	\label{fig:method}
\end{figure}

We now describe RelICL, which avoids the drawbacks of feature explosion and
interaction blindness. At its heart, RelICL is a very simple approach; see
\cref{fig:method}. Its key idea is to use a TFM for both final predictions (as
in DFS) but also for feature extraction and feature fusion (unlike DFS). More
specifically, RelICL represents context-table rows via row embeddings extracted
by the TFM. It fuses information from a referenced context table by adding
pooled row embeddings as additional features. Since the number of added features
depends only on the number of directly referenced tables (i.e., the local
neighborhood), feature explosion is avoided. Since row embeddings capture
feature interactions, interaction blindness is avoided as well.

We first discuss the assumptions underlying RelICL, and then describe the
approach in more detail.

\subsection{Row Embeddings are Good Features}

We consider TFMs $\tfm$ based on row embeddings. To obtain predictions for a
table $R$, $\tfm$ first uses an encoder $\tenc$ to extract row embeddings of
the training and query rows\footnote{The row embedding of a query row $r$
	depends only on $r$ and the entire set of train rows. It does not depend on
	any other query rows.} (without the target attribute) and subsequently applies
a prediction head $\tpred$ (using the target attribute of the train rows). In
short, $\tfm(R)=\tpred(\tenc(R))$. Most most recent TFMs, including TabICL
\citep{Qu2026}, TabPFN \citep{Hollmann2025a}, TabDPT \citep{Ma2025}, LimiX
\citep{Zhang2025}, and TabFM~\citep{Kong2026}, follow this approach.

\begin{assumption}[Row embeddings are good features]
	\label{assm:embedding-sufficiency}
	The performance of $\tfm(\tenc(R))$ matches the one of $\tfm(R)$.
\end{assumption}

This means that we can apply the encoder multiple times without loss in
performance (since $\tfm(\tenc(R))\approx \tpred(\tenc(\tenc(R)))$). This
assumption is crucial to RelICL, because it allows us to replace the features of
a table $R$ by row embeddings without loss in performance. This in turn
simplifies RelICL's feature fusion: use (pooled) row embeddings from context
tables as additional features. Intuitively, this assumption appears reasonable:
since $\tpred$ only relies on $\tenc(R)$ (but not on $R$ itself), the row
embeddings must summarize the corresponding rows of $R$ adequately for  $\tfm$
to work well.

We tested this assumption on ten single-table binary classification datasets
from OpenML \citep{Vanschoren2014}. For each task, we compared TabICL on the
original table ($\tfm(R)$) against TabICL on the row embeddings
($\tfm(\tenc(R))$). \Cref{app:early-fusion} provides dataset statistics and
further details, \Cref{tab:early-fusion} shows the results. We found that for
most datasets (all except philippine and jasmine), there was no statistically
significant loss in performance, providing some justification for our
assumption.

\begin{table}
	\caption{%
		Replacing the feature columns by a row embedding (test AUROC (\%), higher
		is better, mean $\pm$ SEM over three seeds).
		$\tfm(R)$ runs TabICL on the table $R$.
		$\tfm(\tenc(R))$ replaces all feature columns by TabICL's respective row
		embedding.
		$\Delta$ is the paired difference against the original data of the same
		seed.
		The last row averages the ten datasets, so its SEM is taken across
		datasets rather than across seeds.
	}
	\label{tab:early-fusion}
	\centering
	\footnotesize
	\begin{tabular}{lccc}
\toprule
 & \textbf{$\tfm(R)$} & \multicolumn{2}{c}{\textbf{$\tfm(\tenc(R))$}} \\
\cmidrule(lr){2-2}\cmidrule(lr){3-4}
\textbf{Dataset} & \textbf{AUROC} & \textbf{AUROC} & \textbf{$\Delta$} \\
\midrule
bioresponse & 87.7 $\pm$ 1.0 & 87.5 $\pm$ 0.8 & -0.3 $\pm$ 0.3 \\
churn & 92.8 $\pm$ 0.8 & 92.6 $\pm$ 0.6 & -0.2 $\pm$ 0.2 \\
compass & 72.3 $\pm$ 0.7 & 72.4 $\pm$ 0.7 & +0.1 $\pm$ 0.1 \\
credit-g & 82.2 $\pm$ 1.8 & 83.1 $\pm$ 0.4 & +0.9 $\pm$ 1.4 \\
jasmine & 89.8 $\pm$ 1.1 & 86.8 $\pm$ 0.4 & -3.0 $\pm$ 0.6 \\
kr-vs-kp & 100.0 $\pm$ 0.0 & 100.0 $\pm$ 0.0 & 0.0 $\pm$ 0.0 \\
philippine & 93.6 $\pm$ 0.1 & 84.7 $\pm$ 0.6 & -8.9 $\pm$ 0.6 \\
phoneme & 97.6 $\pm$ 0.0 & 96.2 $\pm$ 0.2 & -1.5 $\pm$ 0.2 \\
qsar-biodeg & 92.4 $\pm$ 2.1 & 91.3 $\pm$ 2.4 & -1.1 $\pm$ 0.4 \\
spambase & 99.1 $\pm$ 0.0 & 98.8 $\pm$ 0.0 & -0.4 $\pm$ 0.0 \\
\midrule
\textbf{$\emptyset$} & 90.8 $\pm$ 2.7 & 89.3 $\pm$ 2.6 & -1.4 $\pm$ 0.9 \\
\bottomrule
\end{tabular}

\end{table}

\subsection{Feature Extraction and Feature Fusion}
\label{sec:extraction}

RelICL operates in three phases:
\begin{enumerate}
	\item Outward traversal to determine relevant context table rows
	\item Inward traversal to extract and fuse features from context tables
	\item Prediction on the extended task table
\end{enumerate}

The first two phases rely on a partial ordering of the tables in the database,
which we represent as an \emph{inference DAG} (see \cref{fig:method-dag}) with
root $T$ (the task table). An edge $R\to S$ in the inference DAG states that (i)
there is a reference between table $R$ and table $S$ and (ii) the rows of $R$
should be fused into the rows of $S$. To obtain the inference DAG, RelICL uses
breadth-first search starting at the task table so that tables more distant to
the task table are fused into tables closer to the task table.

\textbf{Outward traversal.} Not every row in every context table is
\emph{relevant} in that it is directly or indirectly referenced by a row in the
task table. Irrelevant rows do not need to be propagated, and the goal of the
outward traversal is to determine and prune such rows. In
\cref{fig:method-input}, for instance, there is no prediction task for driver 3
so that the results of driver 3 are irrelevant. Pruning irrelevant rows is
especially beneficial due to the asymptotically quadratic cost of attention
\citep{Vaswani2017} with respect to the number of rows in typical tabular TFMs
\citep{Qu2026,Hollmann2025a}.

During outward traversal, we traverse the inference DAG in reverse topological
order, starting at the task table. When processing a table, we first determine
its set of relevant rows (all for the task table) and then propagate the
corresponding set of relevant keys along every reversed edge (e.g., $\sset{1,2}$
from \emph{task}$\to$\emph{results} or $\sset{11{-}14,15,16,\ldots}$ from
\emph{results}$\to$\emph{lap times}). All rows that do not contain a relevant
key are marked irrelevant and dropped (e.g., row $19$ in
\emph{results}).

\textbf{Inward traversal.} The goal of inward traversal is to propagate
information from the context tables towards the task table, ultimately
constructing the extended task table. This time, the inference DAG is processed
in topological order. Whenever a table $R$ is processed, we fuse information
from its children (along incoming edges) into the table to obtain $R^+$, extract
and pool row embeddings of $R^+$, and finally propagate information to the
parents (along outgoing edges). Each row $r^+$ of $R^+$ is meant to represent (i)
the corresponding row $r$ from $R$ and (ii) all the rows from \emph{all
	descendant tables} referenced by $r$.

\textbf{Feature extraction.} The extraction of features for some table $R^+$
makes use of \cref{assm:embedding-sufficiency}: we run the encoder $\tenc$ on
$R^+$ to obtain a row embedding for every row of $R^+$ (cf.~\cref{fig:method-fusion}). In contrast to using the rows
directly, this has the advantage that (i) the size of the row embeddings is
independent of the width of the table and (ii) each element of each row
embedding depends on the entire row (and not just one column). The first
advantage allows us to avoid feature explosion: adding features to $R$ to obtain
$R^+$ does not affect the size of the information propagated towards $R$'s
parents. The second advantage avoids interaction blindness.

\textbf{Feature fusion.} We now describe how to fuse information from $R$'s
children into $R$ to obtain $R^+$. Consider a child $C\to R$ in the inference DAG.
Since tables are processed in topological order, $C$ has already been processed
when processing $R$, so that an embedding for every row of $C^+$ is available. We
now extend $R$ by a block of columns holding pooled row embeddings. In more
detail, for every row $r$ in $R$, we pool the row embeddings of the rows in $C$
that are referenced\footnote{For relational deep learning task, pooling is
	temporal: we exclude any ``future rows'' from pooling. This is what is done in
	\cref{fig:method-fusion} to fuse \emph{results}$^+$ into \emph{task}.} by $r$
(e.g., row $11$ of \emph{results} references the first two rows of \emph{lap
	times}), e.g., using sum, average, or max pooling. Since the latter two
methods are oblivious to the number of referenced rows, we also add an explicit
\emph{count column} with that number in the spirit of the degree scalers of
\citet{Corso2020}. These numbers are shown in parentheses in
\cref{fig:method-fusion}.

\textbf{Prediction.} The inward traversal stops after processing the task table
$T$ to obtain the extended task table $T^+$. The number of features in $T^+$ is
determined by the number of its children in the inference DAG (one in
\cref{fig:method}), not the number of tables in the entire database. RelICL does
not need (nor does it use) a depth bound to avoid feature explosion. A row $t^+\in
	T^+$ thus summarizes all the rows referenced from the corresponding row $t\in T$,
no matter how far away. To obtain the final prediction, RelICL runs the entire
tabular model on the extended task table, i.e., $\tfm(T^+)$.

\section{Experiments}
\label{sec:experiments}

We performed an experimental study on RelBench in which we compared RelICL to
the state-of-the-art DFS-based approaches and other methods pretrained on
relational data. We found that RelICL (in its default configuration) was on par
with the strongest DFS-based approaches and outperformed the pretrained
relational approaches. We also provide evidence that the underlying TFM used by
RelICL can be easily swapped. Our complete source code and experimental results
are available
at \url{https://github.com/uma-pi1/relicl}.

\subsection{Experimental setup}
\label{sec:exp-setup}

\textbf{Data.} We used the 21 entity tasks from RelBench v2 \citep{Gu2026a}.
They span seven different relational databases in application areas such as
e-commerce and retail, social and online data, racing, and clinical data. The
databases vary in size from under 100\,k to over 44\,M rows. The benchmark
consists of 12 binary classification tasks (e.g., whether a customer churns or
the outcome of a clinical trial) and 9 regression tasks (e.g., a driver's
finishing position or an item's future sales). \cref{app:datasets} reports
database and task statistics.

\textbf{Baselines.} We compared RelICL against four methods, two of them based
on DFS and two of them pretrained on relational data. The DFS-based methods are
RDBLearn \citep{Xu2026a} and TabPFN-Rel \citep{Hayler2026}, which extends
RDBLearn with feature engineering (e.g., cyclic calendar features or
temporal-lag features).\footnote{%
	We consider the use of such features orthogonal to our work. It is
	conceivable, however, that they would also be beneficial for RelICL.
}
The pretrained relational approaches are Relational Transformers (RT)
\citep{Ranjan2025a} and PluRel \citep{Kothapalli2026}. Note we exclude trained
models (such as RelGNN \citep{Chen2025} or Rel-GT \citep{Dwivedi2025b}) or
hosted, closed-source models (such as the API version of TabPFN-Rel and KumoRFM-2
\citep{Hudovernik2026a}) from our study. The results of these models are given
\cref{app:all-models} for completeness.

\textbf{Metrics.} We report AUROC for classification tasks and MAE as well as
$R^2$ for regression tasks, each on the respective RelBench test sets. For prior
models, we report the results from the respective papers. Note that some papers
did not publish both MAE and $R^2$, i.e., we cannot report all metrics.

\textbf{RelICL implementation.} RelICL uses a number of implementation tricks to
speed up processing. First, it uses dimensionality reduction (e.g., random
projections) to reduce the size of the row embeddings and hence inference costs.
\Cref{app:early-fusion} provides evidence that this reduction barely affects
performance. Second, it sub-samples large tables to a fixed maximum size.
Finally, it merges the tables of some many-to-one relationships, which roughly
corresponds to propagating rows instead of row embeddings when no pooling is
needed. \cref{app:schema-rewrites} contains more details.

\begin{table}
	\caption{%
		Head-to-head wins over all regression and classification tasks combined. The
		baselines report different subsets of the tasks and metrics, so an average
		over all tasks would score each model on different tasks. We therefore
		compare the models pair by pair, only on the tasks both report. A cell
		counts the tasks the row model won over the column model divided by the
		total number of shared tasks. A pair shares a task only where both models
		report a common metric for it. RT and PluRel share no regression task with
		RDBLearn and TabPFN-Rel. Bold marks the model that won the pair.
		\Cref{tab:regression-main,tab:classification-main} provide the more details.}
	\label{tab:win-matrix-combined}
	\centering
	\footnotesize
	\begin{threeparttable}
		\begin{tabular}{lccccc}
\toprule
 & \multicolumn{2}{c}{\textbf{Pretrained (rel.)}} & \multicolumn{3}{c}{\textbf{Pretrained (tab.)}} \\
\cmidrule(lr){2-3}\cmidrule(lr){4-6}
\textbf{Model} & \textbf{RT} & \textbf{PluRel} & \textbf{RDBLearn} & \textbf{TabPFN-Rel} & \textbf{RelICL} \\
\midrule
RT & -- & \phantom{0}6\,/\,18\rlap{$^{*}$} & \phantom{0}0\,/\,8\phantom{0} & \phantom{0}2\,/\,10 & \phantom{0}7\,/\,18\rlap{$^{*}$} \\
PluRel & \textbf{11\,/\,18}\rlap{$^{*}$} & -- & \phantom{0}0\,/\,8\phantom{0} & \phantom{0}2\,/\,10 & \phantom{0}5\,/\,18\rlap{$^{*}$} \\
RDBLearn & \textbf{\phantom{0}8\,/\,8\phantom{0}} & \textbf{\phantom{0}8\,/\,8\phantom{0}} & -- & \phantom{0}1\,/\,15 & \phantom{0}5\,/\,15 \\
TabPFN-Rel & \textbf{\phantom{0}8\,/\,10} & \textbf{\phantom{0}8\,/\,10} & \textbf{13\,/\,15} & -- & 10\,/\,21 \\
RelICL & \textbf{11\,/\,18}\rlap{$^{*}$} & \textbf{13\,/\,18}\rlap{$^{*}$} & \textbf{10\,/\,15} & \textbf{11\,/\,21} & -- \\
\bottomrule
\end{tabular}

		\begin{tablenotes}
			\item[*] \mbox{MAE results were not available for all tasks. In these cases $R^2$ was used.}
		\end{tablenotes}
	\end{threeparttable}
\end{table}

\textbf{Hyperparameters.} We ran RelICL in its default configuration throughout.
This configuration ensembles RelICL predictions using one global hyperparameter
setting with one (regression) and three (classification) additional RelICL predictions,
each using a hyperparameter setting differing in exactly one hyperparameter. The
same ensemble members are used for all classification and regression tasks,
respectively. More details are given in \cref{app:ensembling}.

RelICL has 11 hyperparameters (such as sample sizes or the amount of
dimensionality reduction). The global hyperparameter settings have been selected
via random search over 100 trials per task, optimizing for overall validation AUROC
(classification) and MAE (regression) on a subset of 6 classification and 6
regression tasks from RelBench. The ensemble members differ in hyperparameters
that are highly-sensitive according to this search. \cref{app:hpo} has further
details.

\textbf{Backbones.} RelICL by default used TabICL v2~\citep{Qu2026} as its
backbone TFM. We also report results with TabPFN v3~\citep{Hollmann2025a}
using the same default configuration.

\begin{table}[!t]
	\centering
	\footnotesize
	\sbox0{\scalebox{0.94}{\begin{tabular}{lr@{\;/\;}lr@{\;/\;}lr@{\;/\;}lr@{\;/\;}lr@{\;/\;}l}
\toprule
 & \multicolumn{4}{c}{\textbf{Pretrained (rel.)}} & \multicolumn{6}{c}{\textbf{Pretrained (tab.)}} \\
\cmidrule(lr){2-5}\cmidrule(lr){6-11}
\textbf{Task} & \multicolumn{2}{c}{\textbf{RT}} & \multicolumn{2}{c}{\textbf{PluRel}} & \multicolumn{2}{c}{\textbf{RDBLearn}} & \multicolumn{2}{c}{\textbf{TabPFN-Rel}} & \multicolumn{2}{c}{\textbf{RelICL}} \\
\midrule
item-ltv & -- & 32.5 & -- & \textbf{40.5} & 48.56 & -- & 47.80 & -- & \textbf{43.97} & 28.6 \\
user-ltv & -- & \textbf{36.9} & -- & 18.5 & 14.54 & -- & \textbf{14.40} & -- & 14.78 & 21.0 \\
ad-ctr & -- & 4.5 & -- & 4.9 & \textbf{0.03} & -- & \textbf{0.03} & -- & 0.04 & \textbf{18.3} \\
user-attendance & \multicolumn{2}{c}{--} & \multicolumn{2}{c}{--} & \multicolumn{2}{c}{--} & \textbf{0.24} & -- & \textbf{0.24} & \textbf{5.1} \\
driver-position & -- & 52.4 & -- & \textbf{55.5} & \multicolumn{2}{c}{--} & 3.76 & -- & \textbf{3.68} & 25.2 \\
item-sales & -- & 14.0 & -- & 20.0 & 0.06 & -- & 0.06 & -- & \textbf{0.04} & \textbf{69.2} \\
post-votes & -- & \textbf{33.9} & -- & 25.5 & 0.07 & -- & 0.07 & -- & \textbf{0.06} & 25.6 \\
site-success & -- & 4.5 & -- & \textbf{38.6} & 0.42 & -- & \textbf{0.39} & -- & 0.40 & -22.0 \\
study-adverse & -- & 2.6 & -- & 1.6 & 43.91 & -- & 42.60 & -- & \textbf{32.42} & \textbf{54.0} \\
\bottomrule
\end{tabular}
}}%
	\sbox2{\scalebox{0.94}{\begin{tabular}{c}
\toprule
\textbf{Pretrained (tab.)} \\
\cmidrule(lr){1-1}
\textbf{RelICL (TabPFN)} \\
\midrule
43.41 / 29.2 \\
15.48 / 16.5 \\
\phantom{0}0.05 / 0.1\phantom{0} \\
\phantom{0}0.30 / 17.2 \\
\phantom{0}4.00 / 13.4 \\
\phantom{0}0.04 / 50.1 \\
\phantom{0}0.08 / 17.0 \\
\phantom{0}0.43 / -7.1 \\
29.97 / 56.9 \\
\bottomrule
\end{tabular}
}}%
	\begin{tabular}{@{}c@{\hspace{1.5em}}c@{}}
		\begin{minipage}[t]{\wd0}
			\caption{%
				Regression, MAE (lower is better) / $R^2$ (\%, higher is better).
				The table gives the per-task numbers behind the regression part of
				\cref{tab:win-matrix-combined}.
				It shows both metrics because RT and PluRel report only $R^2$, while
				RDBLearn and TabPFN-Rel report only MAE.
				RelICL reports both, so it can be compared with either group.
				Bold marks the best value per task and metric.
			}
			\label{tab:regression-main}
		\end{minipage} &
		\begin{minipage}[t]{\wd2}
			\caption{%
				RelICL with the TabPFN backbone.
			}
			\label{tab:regression-tabpfn}
		\end{minipage}                                        \\
		\usebox0                                                             & \usebox2 \\
		\multicolumn{2}{@{}l@{}}{-- Not reported.}
	\end{tabular}
\end{table}

\begin{table}[!t]
	\centering
	\footnotesize
	\sbox0{\scalebox{0.94}{\begin{tabular}{lccccc}
\toprule
 & \multicolumn{2}{c}{\textbf{Pretrained (rel.)}} & \multicolumn{3}{c}{\textbf{Pretrained (tab.)}} \\
\cmidrule(lr){2-3}\cmidrule(lr){4-6}
\textbf{Task} & \textbf{RT} & \textbf{PluRel} & \textbf{RDBLearn} & \textbf{TabPFN-Rel} & \textbf{RelICL} \\
\midrule
item-churn & 70.9 & 72.5 & 82.1 & \textbf{82.8} & 78.5 \\
user-churn (ama) & 64.0 & 65.0 & 67.6 & \textbf{70.2} & 65.4 \\
user-clicks & 59.5 & 47.9 & \textbf{69.0} & 61.5 & 64.3 \\
user-visits & 61.8 & 63.4 & 65.5 & \textbf{66.9} & 66.2 \\
user-ignore & -- & -- & -- & 70.1 & \textbf{88.5} \\
user-repeat & -- & -- & -- & 76.9 & \textbf{79.3} \\
driver-dnf & \textbf{81.2} & 81.0 & -- & 71.5 & 73.6 \\
driver-top3 & \textbf{89.3} & 88.4 & -- & 79.3 & 81.4 \\
user-churn (hm) & 62.8 & 66.0 & 68.0 & \textbf{70.6} & 68.4 \\
user-badge & 80.1 & 82.0 & 85.3 & \textbf{86.4} & 86.3 \\
user-engagement & 75.7 & 86.2 & 89.4 & \textbf{90.6} & 90.0 \\
study-outcome & 51.8 & 51.8 & 71.6 & 73.1 & \textbf{73.4} \\
\bottomrule
\end{tabular}
}}%
	\sbox2{\scalebox{0.94}{\begin{tabular}{c}
\toprule
\textbf{Pretrained (tab.)} \\
\cmidrule(lr){1-1}
\textbf{RelICL (TabPFN)} \\
\midrule
78.4 \\
64.6 \\
58.7 \\
65.7 \\
87.9 \\
77.7 \\
74.2 \\
78.4 \\
67.8 \\
84.9 \\
89.5 \\
73.6 \\
\bottomrule
\end{tabular}
}}%
	\begin{tabular}{@{}c@{\hspace{1.5em}}c@{}}
		\begin{minipage}[t]{\wd0}
			\caption{%
				Classification (AUROC (\%), higher is better).
				The table gives the per-task numbers behind the classification part of
				\cref{tab:win-matrix-combined}.
				Bold marks the best value per task.
			}
			\label{tab:classification-main}
		\end{minipage} &
		\begin{minipage}[t]{\wd2}
			\caption{RelICL with the  TabPFN backbone.}
			\label{tab:classification-tabpfn}
		\end{minipage}                                        \\
		\usebox0                                                               & \usebox2 \\
		\multicolumn{2}{@{}l@{}}{-- Not reported.}
	\end{tabular}
\end{table}

\subsection{Results}
\label{sec:exp-results}

\textbf{Main results.} \Cref{tab:win-matrix-combined} compares the models pair
by pair over the tasks each pair shares, regression and classification combined.
More details are provided in \Cref{tab:regression-main,tab:classification-main}
which report performance numbers per task type. RelICL either outperformed or
was on par with all models. The closest comparison was with TabPFN-Rel, which
was published concurrently to this work. TabPFN-Rel's feature engineering
clearly boosted its performance in comparison to RDBLearn (and could potentially
also boost RelICL's performance). Against the models pretrained on relational
data, RelICL outperformed RT and PluRel without relational pretraining, mostly
on classification, despite not being tuned for the $R^2$ metric on which these
models have been tuned.

\textbf{Backbones.} \Cref{tab:regression-tabpfn,tab:classification-tabpfn}
report RelICL's results with TabPFN v3 (in the same default configuration).
TabICL v2 was ahead on most tasks of both types, although the margin was small
for classification. RelICL thus did not depend on TabICL specifically.

\textbf{Limitations.} Our experimental study only considered classification and
regression tasks from RelBench.
We did not reproduce baseline numbers, but reported numbers from the respective
papers, which may differ in tuning, seeds, and evaluation protocol, and
uncertainties are commonly not reported. A win in \cref{tab:win-matrix-combined}
may thus come from a performance gap that is not statistically significant.
\cref{assm:embedding-sufficiency} does not hold on every table: replacing the
features by the row embedding cost 8.9 AUROC points on philippine and 3.0 on
jasmine (\cref{app:early-fusion}). Finally, the synthetic tasks of
\cref{sec:dfs} show that limited depth and interaction blindness are indeed
weaknesses of DFS-based approaches, whereas the RelBench results indicate that
DFS models may nevertheless perform well.

\section{Conclusion}
\label{sec:conclusion}

Feature explosion and interaction blindness are two inherent limitations of deep
feature synthesis for relational learning with tabular foundation models. We
explored whether these limitations can be avoided while keeping its
benefits. We proposed RelICL, a simple training-free approach that uses
a tabular foundation model for multiple purposes (feature extraction, feature
fusion, and final prediction). RelICL does not suffer from the limitations of
DFS and had strong empirical performance in our experimental study.

\clearpage
\bibliography{iclr2027_conference}

\begin{thebibliography}{61}
\providecommand{\natexlab}[1]{#1}
\providecommand{\url}[1]{\texttt{#1}}
\expandafter\ifx\csname urlstyle\endcsname\relax
  \providecommand{\doi}[1]{doi: #1}\else
  \providecommand{\doi}{doi: \begingroup \urlstyle{rm}\Url}\fi

\bibitem[Bardenet et~al.(2013)Bardenet, Brendel, K{\'e}gl, and Sebag]{Bardenet2013}
R{\'e}mi Bardenet, M{\'a}ty{\'a}s Brendel, Bal{\'a}zs K{\'e}gl, and Michele Sebag.
\newblock Collaborative hyperparameter tuning.
\newblock In \emph{International Conference on Machine Learning}, pp.\  199--207. PMLR, 2013.

\bibitem[Chen et~al.(2025)Chen, Kanatsoulis, and Leskovec]{Chen2025}
Tianlang Chen, Charilaos Kanatsoulis, and Jure Leskovec.
\newblock {{RelGNN}}: {{Composite Message Passing}} for {{Relational Deep Learning}}.
\newblock In \emph{Forty-Second {{International Conference}} on {{Machine Learning}}}, June 2025.

\bibitem[Chen \& Guestrin(2016)Chen and Guestrin]{Chen2016}
Tianqi Chen and Carlos Guestrin.
\newblock {{XGBoost}}: {{A Scalable Tree Boosting System}}.
\newblock In \emph{Proceedings of the 22nd {{ACM SIGKDD International Conference}} on {{Knowledge Discovery}} and {{Data Mining}}}, pp.\  785--794, San Francisco California USA, August 2016. ACM.
\newblock ISBN 978-1-4503-4232-2.
\newblock \doi{10.1145/2939672.2939785}.

\bibitem[Codd(1970)]{Codd1970}
E.~F. Codd.
\newblock A relational model of data for large shared data banks.
\newblock \emph{Commun. ACM}, 13\penalty0 (6):\penalty0 377--387, June 1970.
\newblock ISSN 0001-0782, 1557-7317.
\newblock \doi{10.1145/362384.362685}.

\bibitem[Corso et~al.(2020)Corso, Cavalleri, Beaini, Li{\`o}, and Veli{\v c}kovi{\'c}]{Corso2020}
Gabriele Corso, Luca Cavalleri, Dominique Beaini, Pietro Li{\`o}, and Petar Veli{\v c}kovi{\'c}.
\newblock Principal neighbourhood aggregation for graph nets.
\newblock \emph{Advances in neural information processing systems}, 33:\penalty0 13260--13271, 2020.

\bibitem[Cvitkovic(2020)]{Cvitkovic2020}
Milan Cvitkovic.
\newblock Supervised {{Learning}} on {{Relational Databases}} with {{Graph Neural Networks}}, February 2020.

\bibitem[Dasgupta \& Gupta(2003)Dasgupta and Gupta]{Dasgupta2003}
Sanjoy Dasgupta and Anupam Gupta.
\newblock An elementary proof of a theorem of {{Johnson}} and {{Lindenstrauss}}.
\newblock \emph{Random Structures \& Algorithms}, 22\penalty0 (1):\penalty0 60--65, 2003.
\newblock ISSN 1098-2418.
\newblock \doi{10.1002/rsa.10073}.

\bibitem[Dem{\v s}ar(2006)]{Demsar2006}
Janez Dem{\v s}ar.
\newblock Statistical comparisons of classifiers over multiple data sets.
\newblock \emph{Journal of Machine learning research}, 7\penalty0 (Jan):\penalty0 1--30, 2006.

\bibitem[Dwivedi \& Bresson(2021)Dwivedi and Bresson]{Dwivedi2021a}
Vijay~Prakash Dwivedi and Xavier Bresson.
\newblock A {{Generalization}} of {{Transformer Networks}} to {{Graphs}}.
\newblock \emph{AAAI Workshop on Deep Learning on Graphs: Methods and Applications}, January 2021.

\bibitem[Dwivedi et~al.(2025)Dwivedi, Jaladi, Shen, Lopez, Kanatsoulis, Puri, Fey, and Leskovec]{Dwivedi2025b}
Vijay~Prakash Dwivedi, Sri Jaladi, Yangyi Shen, Federico Lopez, Charilaos~I. Kanatsoulis, Rishi Puri, Matthias Fey, and Jure Leskovec.
\newblock Relational {{Graph Transformer}}.
\newblock In \emph{The {{Fourteenth International Conference}} on {{Learning Representations}}}, October 2025.

\bibitem[Erickson et~al.(2020)Erickson, Mueller, Shirkov, Zhang, Larroy, Li, and Smola]{Erickson2020a}
Nick Erickson, Jonas Mueller, Alexander Shirkov, Hang Zhang, Pedro Larroy, Mu~Li, and Alexander Smola.
\newblock {{AutoGluon-Tabular}}: {{Robust}} and {{Accurate AutoML}} for {{Structured Data}}, March 2020.

\bibitem[Erickson et~al.(2025)Erickson, Purucker, Tschalzev, Holzm{\"u}ller, Desai, Salinas, and Hutter]{Erickson2025}
Nick Erickson, Lennart Purucker, Andrej Tschalzev, David Holzm{\"u}ller, Prateek~Mutalik Desai, David Salinas, and Frank Hutter.
\newblock {{TabArena}}: {{A Living Benchmark}} for {{Machine Learning}} on {{Tabular Data}}, June 2025.

\bibitem[Fan et~al.(2010)Fan, Zhong, Peng, Verscheure, Zhang, Ren, Yan, and Yang]{Fan2010}
Wei Fan, Erheng Zhong, Jing Peng, Olivier Verscheure, Kun Zhang, Jiangtao Ren, Rong Yan, and Qiang Yang.
\newblock Generalized and {{Heuristic-Free Feature Construction}} for {{Improved Accuracy}}.
\newblock In \emph{Proceedings of the 2010 {{SIAM International Conference}} on {{Data Mining}}}, pp.\  629--640. {Society for Industrial and Applied Mathematics}, April 2010.
\newblock ISBN 978-0-89871-703-7 978-1-61197-280-1.
\newblock \doi{10.1137/1.9781611972801.55}.

\bibitem[Ferrini et~al.(2024)Ferrini, Longa, Passerini, and Jaeger]{Ferrini2024}
Francesco Ferrini, Antonio Longa, Andrea Passerini, and Manfred Jaeger.
\newblock Meta-{{Path Learning}} for {{Multi-Relational Graph Neural Networks}}.
\newblock In \emph{Proceedings of the {{Second Learning}} on {{Graphs Conference}}}, pp.\  2:1--2:17. PMLR, April 2024.

\bibitem[Fey et~al.(2024)Fey, Hu, Huang, Lenssen, Ranjan, Robinson, Ying, You, and Leskovec]{Fey2024}
Matthias Fey, Weihua Hu, Kexin Huang, Jan~Eric Lenssen, Rishabh Ranjan, Joshua Robinson, Rex Ying, Jiaxuan You, and Jure Leskovec.
\newblock Position: {{Relational Deep Learning}} - {{Graph Representation Learning}} on {{Relational Databases}}.
\newblock In \emph{Proceedings of the 41st {{International Conference}} on {{Machine Learning}}}, pp.\  13592--13607. PMLR, July 2024.

\bibitem[Fey et~al.(2025)Fey, Kocijan, Lopez, Lenssen, and Leskovec]{Fey2025a}
Matthias Fey, Vid Kocijan, Federico Lopez, J.~Lenssen, and Jure Leskovec.
\newblock Kumorfm: {{A}} foundation model for in-context learning on relational data.
\newblock \emph{Technical whitepaper, Kumo AI}, 2025.

\bibitem[{Garcia-Molina} et~al.(2009){Garcia-Molina}, Ullman, and Widom]{Garcia-Molina2009}
Hector {Garcia-Molina}, Jeffrey Ullman, and Jennifer Widom.
\newblock \emph{Database {{Systems}}: {{The Complete Book}}}.
\newblock Pearson, Upper Saddle River, N.J, 2009.
\newblock ISBN 978-0-13-187325-4.

\bibitem[Gardner et~al.(2024)Gardner, Perdomo, and Schmidt]{Gardner2024}
Josh Gardner, Juan~C. Perdomo, and Ludwig Schmidt.
\newblock Large scale transfer learning for tabular data via language modeling.
\newblock \emph{Advances in Neural Information Processing Systems}, 37:\penalty0 45155--45205, 2024.

\bibitem[Gorishniy et~al.(2021)Gorishniy, Rubachev, Khrulkov, and Babenko]{Gorishniy2021}
Yury Gorishniy, Ivan Rubachev, Valentin Khrulkov, and Artem Babenko.
\newblock Revisiting {{Deep Learning Models}} for {{Tabular Data}}.
\newblock In \emph{Advances in {{Neural Information Processing Systems}}}, November 2021.

\bibitem[Gu et~al.(2026)Gu, Ranjan, Kanatsoulis, Tang, Jurkovi{\v c}, Hudovernik, Znidar, Chaturvedi, Shroff, Li, and Leskovec]{Gu2026a}
Justin Gu, Rishabh Ranjan, Charilaos~I. Kanatsoulis, Haiming Tang, Martin Jurkovi{\v c}, Valter Hudovernik, Mark Znidar, Pranshu Chaturvedi, Parth Shroff, Fengyu Li, and Jure Leskovec.
\newblock {{RelBench}} v2: {{A Large-Scale Benchmark}} and {{Relational Data Repository}}.
\newblock \emph{ICLR 2026 Workshop on Navigating and Addressing Data Problems for Foundation Models}, March 2026.

\bibitem[Hamilton et~al.(2017)Hamilton, Ying, and Leskovec]{Hamilton2017}
Will Hamilton, Zhitao Ying, and Jure Leskovec.
\newblock Inductive {{Representation Learning}} on {{Large Graphs}}.
\newblock In \emph{Advances in {{Neural Information Processing Systems}}}, volume~30. Curran Associates, Inc., 2017.

\bibitem[Hayler et~al.(2026)Hayler, Fl{\"o}ge, Arazi, Ranjan, Leskovec, Birkel, Roof, Garg, Collins, Sidhoum, K{\"u}bler, Guo, Key, Metzen, Grace, Salinas, Cahu, Bing, J{\"a}ger, {\c C}elik, Manium, Monteiro, Robertson, Chen, Kalfon, Pereda, Wehrhahn, Safaric, Schroeder, Grab, Kriuchkova, Cornu, Singer, Erickson, Balazadeh, Salmon, Alessi, Kaya, Jund, Grinsztajn, LeCun, Sch{\"o}lkopf, Hulsebos, Purucker, Gambhir, Hutter, and Hollmann]{Hayler2026}
Adrian Hayler, Klemens Fl{\"o}ge, Alan Arazi, Rishabh Ranjan, Jure Leskovec, Felix Birkel, Brendan Roof, Anurag Garg, Kristina Collins, Lydia Sidhoum, Jonas K{\"u}bler, Siyuan Guo, Oscar Key, Jan~Hendrik Metzen, Rylee Grace, David Salinas, Arthur Cahu, Simon Bing, Benjamin J{\"a}ger, Tuana {\c C}elik, Mihir Manium, Vitor Monteiro, Jake Robertson, Jerry Chen, Eliott Kalfon, Tom{\'a}s Pereda, Lilly Wehrhahn, Dominik Safaric, Tobias Schroeder, Georg Grab, Diana Kriuchkova, Clara Cornu, Philipp Singer, Nick Erickson, Vahid Balazadeh, Marie Salmon, Simone Alessi, K{\"u}r{\c s}at Kaya, Philipp Jund, L{\'e}o Grinsztajn, Yann LeCun, Bernhard Sch{\"o}lkopf, Madelon Hulsebos, Lennart Purucker, Sauraj Gambhir, Frank Hutter, and Noah Hollmann.
\newblock Advancing {{Open}} and {{Reproducible Relational Learning}}: {{RelArena-}}\${$\alpha\$$}, {{TabPFN-Rel}} and {{RPI}}, August 2026.

\bibitem[Hollmann et~al.(2022)Hollmann, M{\"u}ller, Eggensperger, and Hutter]{Hollmann2022}
Noah Hollmann, Samuel M{\"u}ller, Katharina Eggensperger, and Frank Hutter.
\newblock {{TabPFN}}: {{A Transformer That Solves Small Tabular Classification Problems}} in a {{Second}}.
\newblock In \emph{The {{Eleventh International Conference}} on {{Learning Representations}}}, September 2022.

\bibitem[Hollmann et~al.(2025)Hollmann, M{\"u}ller, Purucker, Krishnakumar, K{\"o}rfer, Hoo, Schirrmeister, and Hutter]{Hollmann2025a}
Noah Hollmann, Samuel M{\"u}ller, Lennart Purucker, Arjun Krishnakumar, Max K{\"o}rfer, Shi~Bin Hoo, Robin~Tibor Schirrmeister, and Frank Hutter.
\newblock Accurate predictions on small data with a tabular foundation model.
\newblock \emph{Nature}, 637\penalty0 (8045):\penalty0 319--326, January 2025.
\newblock ISSN 1476-4687.
\newblock \doi{10.1038/s41586-024-08328-6}.

\bibitem[Holzm{\"u}ller et~al.(2024)Holzm{\"u}ller, Grinsztajn, and Steinwart]{Holzmuller2024}
David Holzm{\"u}ller, L{\'e}o Grinsztajn, and Ingo Steinwart.
\newblock Better by default: {{Strong}} pre-tuned {{MLPs}} and boosted trees on tabular data.
\newblock \emph{Advances in Neural Information Processing Systems}, 37:\penalty0 26577--26658, December 2024.

\bibitem[Hudovernik et~al.(2026)Hudovernik, L{\'o}pez, Kocijan, Nitta, Lenssen, Leskovec, and Fey]{Hudovernik2026a}
Valter Hudovernik, Federico L{\'o}pez, Vid Kocijan, Akihiro Nitta, Jan~Eric Lenssen, Jure Leskovec, and Matthias Fey.
\newblock {{KumoRFM-2}}: {{Scaling Foundation Models}} for {{Relational Learning}}, April 2026.

\bibitem[Hutter et~al.(2014)Hutter, Hoos, and {Leyton-Brown}]{Hutter2014a}
Frank Hutter, Holger Hoos, and Kevin {Leyton-Brown}.
\newblock An efficient approach for assessing hyperparameter importance.
\newblock In \emph{International Conference on Machine Learning}, pp.\  754--762. PMLR, 2014.

\bibitem[Johnson \& Lindenstrauss(1984)Johnson and Lindenstrauss]{Johnson1984}
William~B. Johnson and Joram Lindenstrauss.
\newblock Extensions of {{Lipschitz}} mappings into a {{Hilbert}} space.
\newblock \emph{Contemporary mathematics}, 26\penalty0 (189-206):\penalty0 1, 1984.

\bibitem[Kanter \& Veeramachaneni(2015)Kanter and Veeramachaneni]{Kanter2015}
James~Max Kanter and Kalyan Veeramachaneni.
\newblock Deep feature synthesis: {{Towards}} automating data science endeavors.
\newblock In \emph{2015 {{IEEE}} International Conference on Data Science and Advanced Analytics ({{DSAA}})}, pp.\  1--10. IEEE, 2015.

\bibitem[Ke et~al.(2017)Ke, Meng, Finley, Wang, Chen, Ma, Ye, and Liu]{Ke2017}
Guolin Ke, Qi~Meng, Thomas Finley, Taifeng Wang, Wei Chen, Weidong Ma, Qiwei Ye, and Tie-Yan Liu.
\newblock Lightgbm: {{A}} highly efficient gradient boosting decision tree.
\newblock \emph{Advances in neural information processing systems}, 30, 2017.

\bibitem[Kong \& Das(2026)Kong and Das]{Kong2026}
Weihao Kong and Abhimanyu Das.
\newblock Introducing {{TabFM}}: {{A}} zero-shot foundation model for tabular data.
\newblock https://research.google/blog/introducing-tabfm-a-zero-shot-foundation-model-for-tabular-data, September 2026.

\bibitem[Kothapalli et~al.(2026)Kothapalli, Ranjan, Hudovernik, Dwivedi, Hoffart, Guestrin, and Leskovec]{Kothapalli2026}
Vignesh Kothapalli, Rishabh Ranjan, Valter Hudovernik, Vijay~Prakash Dwivedi, Johannes Hoffart, Carlos Guestrin, and Jure Leskovec.
\newblock {{PluRel}}: {{Synthetic Data}} unlocks {{Scaling Laws}} for {{Relational Foundation Models}}, February 2026.

\bibitem[Lam et~al.(2017)Lam, Thiebaut, Sinn, Chen, Mai, and Alkan]{Lam2017}
Hoang~Thanh Lam, Johann-Michael Thiebaut, Mathieu Sinn, Bei Chen, Tiep Mai, and Oznur Alkan.
\newblock One button machine for automating feature engineering in relational databases, June 2017.

\bibitem[Ma et~al.(2025)Ma, Thomas, Hosseinzadeh, Labach, Cresswell, Golestan, Yu, Caterini, and Volkovs]{Ma2025}
Junwei Ma, Valentin Thomas, Rasa Hosseinzadeh, Alex Labach, Jesse Cresswell, Keyvan Golestan, Guangwei Yu, Anthony~L Caterini, and Maks Volkovs.
\newblock {{TabDPT}}: {{Scaling Tabular Foundation Models}} on {{Real Data}}.
\newblock In \emph{Advances in {{Neural Information Processing Systems}}}, volume 38, Main Conference, pp.\  172692--172722. Curran Associates, Inc., 2025.
\newblock \doi{10.52202/085713-5748}.

\bibitem[Markovitch \& Rosenstein(2002)Markovitch and Rosenstein]{Markovitch2002}
Shaul Markovitch and Dan Rosenstein.
\newblock Feature {{Generation Using General Constructor Functions}}.
\newblock \emph{Machine Learning}, 49\penalty0 (1):\penalty0 59--98, October 2002.
\newblock ISSN 0885-6125, 1573-0565.
\newblock \doi{10.1023/A:1014046307775}.

\bibitem[Motl \& Schulte(2025)Motl and Schulte]{Motl2025a}
Jan Motl and Oliver Schulte.
\newblock The {{CTU Prague Relational Learning Repository}}, March 2025.

\bibitem[Pele{\v s}ka \& {\v S}{\'i}r(2026)Pele{\v s}ka and {\v S}{\'i}r]{Peleska2026b}
Jakub Pele{\v s}ka and Gustav {\v S}{\'i}r.
\newblock {{ReDeLEx}}: {{A Framework}} for~{{Relational Deep Learning Exploration}}.
\newblock In Rita~P. Ribeiro, Bernhard Pfahringer, Nathalie Japkowicz, Pedro Larra{\~n}aga, Al{\'i}pio~M. Jorge, Carlos Soares, Pedro~H. Abreu, and Jo{\~a}o Gama (eds.), \emph{Machine {{Learning}} and {{Knowledge Discovery}} in {{Databases}}. {{Research Track}}}, pp.\  438--456, Cham, 2026. Springer Nature Switzerland.
\newblock ISBN 978-3-032-05981-9.
\newblock \doi{10.1007/978-3-032-05981-9_26}.

\bibitem[Pele{\v s}ka \& {{\v s}{\'i}r}(2025)Pele{\v s}ka and {{\v s}{\'i}r}]{Peleska2025}
Jakub Pele{\v s}ka and Gustav {{\v s}{\'i}r}.
\newblock Tabular {{Transformers Meet Relational Databases}}.
\newblock \emph{ACM Transactions on Intelligent Systems and Technology}, 0, 2025.
\newblock \doi{10.1145/3749991}.

\bibitem[Piramuthu \& Sikora(2009)Piramuthu and Sikora]{Piramuthu2009}
Selwyn Piramuthu and Riyaz~T. Sikora.
\newblock Iterative feature construction for improving inductive learning algorithms.
\newblock \emph{Expert Systems with Applications}, 36\penalty0 (2):\penalty0 3401--3406, 2009.

\bibitem[Prokhorenkova et~al.(2018)Prokhorenkova, Gusev, Vorobev, Dorogush, and Gulin]{Prokhorenkova2018}
Liudmila Prokhorenkova, Gleb Gusev, Aleksandr Vorobev, Anna~Veronika Dorogush, and Andrey Gulin.
\newblock {{CatBoost}}: Unbiased boosting with categorical features.
\newblock \emph{Advances in neural information processing systems}, 31, 2018.

\bibitem[Qu et~al.(2025)Qu, Holzm{\"u}ller, Varoquaux, and Morvan]{Qu2025}
Jingang Qu, David Holzm{\"u}ller, Ga{\"e}l Varoquaux, and Marine~Le Morvan.
\newblock {{TabICL}}: {{A Tabular Foundation Model}} for {{In-Context Learning}} on {{Large Data}}.
\newblock In \emph{Forty-Second {{International Conference}} on {{Machine Learning}}}, June 2025.

\bibitem[Qu et~al.(2026)Qu, Holzm{\"u}ller, Varoquaux, and Morvan]{Qu2026}
Jingang Qu, David Holzm{\"u}ller, Ga{\"e}l Varoquaux, and Marine~Le Morvan.
\newblock {{TabICLv2}}: {{A Better}}, {{Faster}}, {{Scalable}}, and {{Open Tabular Foundation Model}}.
\newblock In \emph{Forty-Third {{International Conference}} on {{Machine Learning}}}, June 2026.

\bibitem[Ranjan et~al.(2025)Ranjan, Hudovernik, Znidar, Kanatsoulis, Upendra, Mohammadi, Meyer, Palczewski, Guestrin, and Leskovec]{Ranjan2025a}
Rishabh Ranjan, Valter Hudovernik, Mark Znidar, Charilaos~I. Kanatsoulis, Roshan~Reddy Upendra, Mahmoud Mohammadi, Joe Meyer, Tom Palczewski, Carlos Guestrin, and Jure Leskovec.
\newblock Relational {{Transformer}}: {{Toward Zero-Shot Foundation Models}} for {{Relational Data}}.
\newblock In \emph{The {{Fourteenth International Conference}} on {{Learning Representations}}}, October 2025.

\bibitem[Robinson et~al.(2024)Robinson, Ranjan, Hu, Huang, Han, Dobles, Fey, Lenssen, Yuan, Zhang, He, and Leskovec]{Robinson2024}
Joshua Robinson, Rishabh Ranjan, Weihua Hu, Kexin Huang, Jiaqi Han, Alejandro Dobles, Matthias Fey, Jan~Eric Lenssen, Yiwen Yuan, Zecheng Zhang, Xinwei He, and Jure Leskovec.
\newblock {{RelBench}}: {{A Benchmark}} for {{Deep Learning}} on {{Relational Databases}}.
\newblock In \emph{The {{Thirty-eight Conference}} on {{Neural Information Processing Systems Datasets}} and {{Benchmarks Track}}}, November 2024.

\bibitem[Rossi et~al.(2020)Rossi, Chamberlain, Frasca, Eynard, Monti, and Bronstein]{Rossi2020}
Emanuele Rossi, Ben Chamberlain, Fabrizio Frasca, Davide Eynard, Federico Monti, and Michael Bronstein.
\newblock Temporal {{Graph Networks}} for {{Deep Learning}} on {{Dynamic Graphs}}, October 2020.

\bibitem[Scarselli et~al.(2008)Scarselli, Gori, Tsoi, Hagenbuchner, and Monfardini]{Scarselli2008}
Franco Scarselli, Marco Gori, Ah~Chung Tsoi, Markus Hagenbuchner, and Gabriele Monfardini.
\newblock The graph neural network model.
\newblock \emph{IEEE transactions on neural networks}, 20\penalty0 (1):\penalty0 61--80, 2008.

\bibitem[Schlichtkrull et~al.(2018)Schlichtkrull, Kipf, Bloem, Van~Den~Berg, Titov, and Welling]{Schlichtkrull2018}
Michael Schlichtkrull, Thomas~N. Kipf, Peter Bloem, Rianne Van~Den~Berg, Ivan Titov, and Max Welling.
\newblock Modeling {{Relational Data}} with {{Graph Convolutional Networks}}.
\newblock In Aldo Gangemi, Roberto Navigli, Maria-Esther Vidal, Pascal Hitzler, Rapha{\"e}l Troncy, Laura Hollink, Anna Tordai, and Mehwish Alam (eds.), \emph{The {{Semantic Web}}}, volume 10843, pp.\  593--607. Springer International Publishing, Cham, 2018.
\newblock ISBN 978-3-319-93416-7 978-3-319-93417-4.
\newblock \doi{10.1007/978-3-319-93417-4_38}.

\bibitem[Shi et~al.(2020)Shi, Zhang, Li, Yang, Li, and Zhou]{Shi2020}
Qitao Shi, Ya-Lin Zhang, Longfei Li, Xinxing Yang, Meng Li, and Jun Zhou.
\newblock Safe: {{Scalable}} automatic feature engineering framework for industrial tasks.
\newblock In \emph{2020 {{IEEE}} 36th {{International Conference}} on {{Data Engineering}} ({{ICDE}})}, pp.\  1645--1656. IEEE, 2020.

\bibitem[{\v S}{\'i}r(2021)]{Sir2021}
Gustav {\v S}{\'i}r.
\newblock \emph{Deep {{Learning}} with {{Relational Logic Representations}}}.
\newblock Czech Technical University, 2021.

\bibitem[Spinaci et~al.(2025)Spinaci, Polewczyk, Schambach, and Thelin]{Spinaci2025}
Marco Spinaci, Marek Polewczyk, Maximilian Schambach, and Sam Thelin.
\newblock {{ConTextTab}}: {{A Semantics-Aware Tabular In-Context Learner}}.
\newblock In \emph{The {{Thirty-ninth Annual Conference}} on {{Neural Information Processing Systems}}}, October 2025.

\bibitem[Sun et~al.(2011)Sun, Han, Yan, Yu, and Wu]{Sun2011}
Yizhou Sun, Jiawei Han, Xifeng Yan, Philip~S. Yu, and Tianyi Wu.
\newblock {{PathSim}}: Meta path-based top-{{K}} similarity search in heterogeneous information networks.
\newblock \emph{Proc. VLDB Endow.}, 4\penalty0 (11):\penalty0 992--1003, August 2011.
\newblock ISSN 2150-8097.
\newblock \doi{10.14778/3402707.3402736}.

\bibitem[Vanschoren et~al.(2014)Vanschoren, Van~Rijn, Bischl, and Torgo]{Vanschoren2014}
Joaquin Vanschoren, Jan~N. Van~Rijn, Bernd Bischl, and Luis Torgo.
\newblock {{OpenML}}: Networked science in machine learning.
\newblock \emph{SIGKDD Explor. Newsl.}, 15\penalty0 (2):\penalty0 49--60, June 2014.
\newblock ISSN 1931-0145, 1931-0153.
\newblock \doi{10.1145/2641190.2641198}.

\bibitem[Vaswani et~al.(2017)Vaswani, Shazeer, Parmar, Uszkoreit, Jones, Gomez, Kaiser, and Polosukhin]{Vaswani2017}
Ashish Vaswani, Noam Shazeer, Niki Parmar, Jakob Uszkoreit, Llion Jones, Aidan~N. Gomez, {\L}ukasz Kaiser, and Illia Polosukhin.
\newblock Attention is all you need.
\newblock In \emph{Proceedings of the 31st {{International Conference}} on {{Neural Information Processing Systems}}}, {{NIPS}}'17, pp.\  6000--6010, Red Hook, NY, USA, December 2017. Curran Associates Inc.
\newblock ISBN 978-1-5108-6096-4.

\bibitem[Wang et~al.(2024)Wang, Gan, Wipf, Cai, Li, Tang, Zhang, Zhang, Mao, Song, Wang, Li, Zhang, Yang, Qin, Lei, Zhang, Zhang, Faloutsos, and Zhang]{Wang2024}
Minjie Wang, Quan Gan, David Wipf, Zhenkun Cai, Ning Li, Jianheng Tang, Yanlin Zhang, Zizhao Zhang, Zunyao Mao, Yakun Song, Yanbo Wang, Jiahang Li, Han Zhang, Guang Yang, Xiao Qin, Chuan Lei, Muhan Zhang, Weinan Zhang, Christos Faloutsos, and Zheng Zhang.
\newblock {{4DBInfer}}: {{A 4D Benchmarking Toolbox}} for {{Graph-Centric Predictive Modeling}} on {{Relational DBs}}, April 2024.

\bibitem[Wang et~al.(2019)Wang, Ji, Shi, Wang, Ye, Cui, and Yu]{Wang2019}
Xiao Wang, Houye Ji, Chuan Shi, Bai Wang, Yanfang Ye, Peng Cui, and Philip~S Yu.
\newblock Heterogeneous {{Graph Attention Network}}.
\newblock In \emph{The {{World Wide Web Conference}}}, {{WWW}} '19, pp.\  2022--2032, New York, NY, USA, May 2019. Association for Computing Machinery.
\newblock ISBN 978-1-4503-6674-8.
\newblock \doi{10.1145/3308558.3313562}.

\bibitem[Wang et~al.(2025)Wang, Wang, Gan, Wang, Yang, Wipf, and Zhang]{Wang2025b}
Yanbo Wang, Xiyuan Wang, Quan Gan, Minjie Wang, Qibin Yang, David Wipf, and Muhan Zhang.
\newblock Griffin: {{Towards}} a {{Graph-Centric Relational Database Foundation Model}}.
\newblock In \emph{Forty-Second {{International Conference}} on {{Machine Learning}}}, June 2025.

\bibitem[Xu et~al.(2026)Xu, Zhang, Gan, Wang, and Wipf]{Xu2026a}
Linjie Xu, Yanlin Zhang, Quan Gan, Minjie Wang, and David Wipf.
\newblock No {{Need}} to {{Train Your RDB Foundation Model}}.
\newblock In \emph{Forty-Third {{International Conference}} on {{Machine Learning}}}, June 2026.

\bibitem[You et~al.(2021)You, {Gomes-Selman}, Ying, and Leskovec]{You2021b}
Jiaxuan You, Jonathan~M. {Gomes-Selman}, Rex Ying, and Jure Leskovec.
\newblock Identity-aware {{Graph Neural Networks}}.
\newblock \emph{Proceedings of the AAAI Conference on Artificial Intelligence}, 35\penalty0 (12):\penalty0 10737--10745, May 2021.
\newblock ISSN 2374-3468.
\newblock \doi{10.1609/aaai.v35i12.17283}.

\bibitem[Zahradn{\'i}k et~al.(2023)Zahradn{\'i}k, Neumann, and {\v S}{\'i}r]{Zahradnik2023}
Luk{\'a}{\v s} Zahradn{\'i}k, Jan Neumann, and Gustav {\v S}{\'i}r.
\newblock A {{Deep Learning Blueprint}} for {{Relational Databases}}.
\newblock In \emph{{{NeurIPS}} 2023 {{Second Table Representation Learning Workshop}}}, October 2023.

\bibitem[Zhang et~al.(2025)Zhang, Ren, Yu, Yuan, Wang, Li, Wu, Mo, Mao, Hao, Dai, Xu, Li, Zhang, He, Wang, Zhang, Xu, Li, Gao, Zou, Liu, Liu, Xu, Cheng, Li, Zhou, Li, Fan, Lin, Han, Li, Lu, Xue, Jiang, Wang, Wang, and Cui]{Zhang2025}
Xingxuan Zhang, Gang Ren, Han Yu, Hao Yuan, Hui Wang, Jiansheng Li, Jiayun Wu, Lang Mo, Li~Mao, Mingchao Hao, Ningbo Dai, Renzhe Xu, Shuyang Li, Tianyang Zhang, Yue He, Yuanrui Wang, Yunjia Zhang, Zijing Xu, Dongzhe Li, Fang Gao, Hao Zou, Jiandong Liu, Jiashuo Liu, Jiawei Xu, Kaijie Cheng, Kehan Li, Linjun Zhou, Qing Li, Shaohua Fan, Xiaoyu Lin, Xinyan Han, Xuanyue Li, Yan Lu, Yuan Xue, Yuanyuan Jiang, Zimu Wang, Zhenlei Wang, and Peng Cui.
\newblock {{LimiX}}: {{Unleashing Structured-Data Modeling Capability}} for {{Generalist Intelligence}}, November 2025.

\bibitem[Zhang et~al.(2026)Zhang, Xu, Gan, Wipf, and Wang]{Zhang2026}
Yanlin Zhang, Linjie Xu, Quan Gan, David Wipf, and Minjie Wang.
\newblock {{RDBLearn}}: {{Simple In-Context Prediction Over Relational Databases}}, February 2026.

\end{thebibliography}
\bibliographystyle{iclr2027_conference}

\clearpage
\appendix

\crefalias{section}{appendix}
\crefalias{subsection}{appendix}

\section{Extended related work}
\label{app:related-work}

This appendix covers further related work which is not directly relevant to
follow the paper.

\textbf{Graph-based approaches.}
Rather than first flattening the schema and then running inference, graph-based
approaches work on the schema directly to make predictions end-to-end.
\citet{Schlichtkrull2018, Cvitkovic2020, Sir2021} were among the first to frame
supervised learning over relational databases as a graph neural network
\citep{Scarselli2008} problem, with one node per row and one edge per foreign
key.
\citet{Fey2024} later formalize and coin this idea as Relational Deep Learning
(RDL).
RDL puts a particular focus on the time domain and hence builds upon temporal
graph learning \citep{Rossi2020}.
In its basic form, a heterogeneous graph neural network such as GraphSAGE
\citep{Hamilton2017} or ID-GNN \citep{You2021b} is trained from scratch for
every database and task \citep{Robinson2024}.
Work since then follows three lines.
The first improves message passing itself.
RelGNN \citep{Chen2025} passes messages across many-to-many relationships in a
single hop and thereby needs fewer message-passing layers.
The second adapts Graph Transformers \citep{Dwivedi2021a} to the relational
domain.
The Relational Graph Transformer \citep{Dwivedi2025b} tokenizes a node by its
type, hop distance, time, and local structure besides its features, i.e., by
properties intrinsic to relational data.
The third pretrains one relational foundation model across databases and applies
it to any database or task, with or without fine-tuning, with the Relational
Transformer \citep{Ranjan2025a}, PluRel \citep{Kothapalli2026}, Griffin
\citep{Wang2025b}, and KumoRFM \citep{Fey2025a, Hudovernik2026a} as examples.
These models attain strong performance on relational benchmarks, but they need
training on relational data, whereas in-context relational prediction asks what
a model reaches without any.
Among graph-based approaches, \citet{Zahradnik2023, Peleska2025} come closest to
our work.
They pass messages over the relational schema with a tabular transformer, but
train it end-to-end per database, whereas RelICL reuses a frozen backbone
without any training.

\textbf{Relational benchmarks.}
Benchmarks for relational data have moved from plain collections of databases to
standardized prediction tasks.
The CTU Prague Relational Learning Repository \citep{Motl2025a} is the largest
such collection, with almost 150 SQL databases.
It proposes a target column for many of them, but no splits or evaluation
protocol.
RelBench \citep{Robinson2024, Gu2026a} adds both and pairs seven databases with
30 tasks.
Each task is an SQL query evaluated at a cutoff time, so a model only sees data
that existed at prediction time.
\citet{Wang2024} release 4DBInfer at the same time.
It varies four dimensions: dataset, task, graph extraction, and base predictor.
We evaluate on RelBench since all baselines we compare against report on it.
Further benchmarks aim to consolidate several approaches into one.
ReDeLEx \citep{Peleska2026b} serves over 70 CTU databases through the RelBench
interface, so their targets become RelBench-style tasks.
Those tasks may be static, whereas RelBench tasks are always temporal.
RelBench v2 \citep{Gu2026a} grows to 11 databases and adds the 4DBInfer tasks.
It also introduces autocomplete tasks, which predict missing values inside a
table.
RelArena \citep{Hayler2026} defines an evaluation protocol on RelBench.
It follows TabArena \citep{Erickson2025} from the tabular domain.

\section{Deep Feature Synthesis and its Limitations}
\label{app:dfs}

\subsection{\Cref*{prop:depth}: Insufficiency of Bounded depth}
\label{app:depth}

\begin{proof}[Proof of \cref{prop:depth}]
	Let the task table reference a table $R_1$, and let each $R_i$ be connected to
	a table $R_{i+1}$ for $i \le d$, every connection a one-to-one reference.
	Only $R_{d+1}$ holds a column besides its keys, namely a uniformly random bit,
	and the target of a task row is that bit.
	Every feature that follows at most $d$ references is then a function of key
	values alone, which are independent of the target.
\end{proof}

\textbf{Data.}
The database holds the task table and a chain of three further tables,
$R_1$, $R_2$, and $R_3$ (\cref{fig:depth}).
The task table references $R_1$, $R_1$ references $R_2$, and $R_2$ references
$R_3$.
Every reference is one-to-one.
The only non-key column in the database is a bit in $R_3$, which equals the
label of the task row.
The label therefore sits three references away from the task table, which is
the construction of \cref{prop:depth} for $d = 2$.
Training and test rows are disjoint, both splits are balanced, and we used
1000 training and 300 test rows.
We ran three seeds for each method.

\begin{figure}
	\centering
	\begingroup
\def\cyT{0}       %
\def\cxT{0}       %
\def\cxA{24}      %
\def\cxB{48}      %
\def\cxC{72}      %

\begin{tikzpicture}[
	x=1mm, y=1mm,
	rounded corners=0.4mm,
	hdr/.style={draw=primary, fill=primary, text=primarytext,
			minimum width=\cellw, minimum height=\cellh, inner sep=0pt,
			font=\tiny, anchor=west},
	dat/.style={hdr, draw=border, fill=white, text=black},
	qry/.style={dat, fill=highlight},
	tname/.style={font=\tiny\itshape, inner sep=0.6pt, anchor=south west},
	fk/.style={border, shorten >=0.4mm, shorten <=0.4mm},
	]
	\reltable{\cxT}{\cyT}{task}
	\relrow{ct}{0}{label/hdr, r1\_id/hdr}
	\relrow{cta}{1}{0/dat, 1/dat}
	\relrow{ctb}{2}{1/dat, 2/dat}
	\relrow{ctc}{3}{{\textbf{?}}/qry, 3/qry}

	\reltable{\cxA}{\cyT}{$R_1$}
	\relrow{ca}{0}{r1\_id/hdr, r2\_id/hdr}
	\relrow{caa}{1}{1/dat, 2/dat}
	\relrow{cab}{2}{2/dat, 1/dat}
	\relrow{cac}{3}{3/dat, 3/dat}

	\reltable{\cxB}{\cyT}{$R_2$}
	\relrow{cb}{0}{r2\_id/hdr, r3\_id/hdr}
	\relrow{cba}{1}{1/dat, 2/dat}
	\relrow{cbb}{2}{2/dat, 1/dat}
	\relrow{cbc}{3}{3/dat, 3/dat}

	\reltable{\cxC}{\cyT}{$R_3$}
	\relrow{cc}{0}{r3\_id/hdr, bit/hdr}
	\relrow{cca}{1}{1/dat, 0/dat}
	\relrow{ccb}{2}{2/dat, 1/dat}
	\relrow{ccc}{3}{3/dat, 1/dat}

	\draw[fk] (ct1.east) -- (ca0.west);
	\draw[fk] (ca1.east) -- (cb0.west);
	\draw[fk] (cb1.east) -- (cc0.west);
\end{tikzpicture}
\endgroup
	\caption{The bounded depth task for three task rows.
		Each line links a key column to the matching column of the next table,
		and every reference is one-to-one.
		The label of a task row equals the bit in $R_3$, three references away.
		The highlighted query row is the one the model has to predict.
	}
	\label{fig:depth}
\end{figure}

\subsection{\Cref*{prop:single-column}: Insufficiency of DFS features}
\label{app:cross-column}

\textbf{Data.}
The database holds the task table and a \emph{signals} table
(\cref{fig:cross-column}).
Training and test entities are disjoint, both splits are balanced, and we used
\numprint{1000} training and 300 test entities.
We ran three seeds for each method, each of which set the random draws of the
data generator and of the model.

\begin{figure}
	\centering
	\begingroup
\def\cyT{0}       %
\def\cxT{0}       %
\def\cxS{24}      %

\begin{tikzpicture}[
	x=1mm, y=1mm,
	rounded corners=0.4mm,
	hdr/.style={draw=primary, fill=primary, text=primarytext,
			minimum width=\cellw, minimum height=\cellh, inner sep=0pt,
			font=\tiny, anchor=west},
	dat/.style={hdr, draw=border, fill=white, text=black},
	qry/.style={dat, fill=highlight},
	tname/.style={font=\tiny\itshape, inner sep=0.6pt, anchor=south west},
	fk/.style={border, shorten >=0.4mm, shorten <=0.4mm},
	]
	\reltable{\cxT}{\cyT}{task}
	\relrow{ct}{0}{entity\_id/hdr, label/hdr}
	\relrow{cta}{1}{1/dat, 1/dat}
	\relrow{ctb}{2}{2/dat, 0/dat}
	\relrow{ctc}{3}{3/qry, {\textbf{?}}/qry}

	\reltable{\cxS}{\cyT}{signals}
	\relrow{cs}{0}{signal\_id/hdr, entity\_id/hdr, a/hdr, b/hdr}
	\relrow{csa}{1}{1/dat, 1/dat, 0/dat, $-1$/dat}
	\relrow{csb}{2}{2/dat, 1/dat, 1/dat, 0/dat}
	\relrow{csc}{3}{3/dat, 1/dat, 2/dat, 1/dat}
	\relrow{csd}{4}{4/dat, 2/dat, 0/dat, 1/dat}
	\relrow{cse}{5}{5/dat, 2/dat, 1/dat, 0/dat}
	\relrow{csf}{6}{6/dat, 2/dat, 2/dat, $-1$/dat}
	\relrow{csg}{7}{7/dat, 3/dat, 0/dat, $-1$/dat}
	\relrow{csh}{8}{8/dat, 3/dat, 1/dat, 0/dat}
	\relrow{csi}{9}{9/dat, 3/dat, 2/dat, 1/dat}

	\draw[fk] (ct0.north) -- ++(0,4) -| (cs1.north);
\end{tikzpicture}
\endgroup
	\caption{
		Showcases a pathological task which cannot be solved through DFS features.
		Entity 1 is positive, so $b$ increases with $a$; entity 2 is negative, so
		$b$ decreases.
		Column $a$ holds $0, 1, 2$ and column $b$ holds $-1, 0, 1$ in both
		classes, so the classes are separated only by which value of $b$ is next
		to which value of $a$.
		Entity 3 is the highlighted query row, whose label the model has to predict
		from its three signal rows.
	}
	\label{fig:cross-column}
\end{figure}

\textbf{Setup and results.}
We ran RDBLearn at its default configuration.
Deep feature synthesis generated nine aggregates over the \emph{signals} table:
the row count, together with the minimum, maximum, mean, and standard deviation
of each of the two columns.
Each of them takes a single distinct value across all training entities,
so the feature generator inside RDBLearn's preprocessor discarded all nine
and the base estimator received only the join key.
RDBLearn therefore can only randomly guess labels.

\section{Method details}
\label{app:method}

\subsection{Schema rewrites}
\label{app:schema-rewrites}

A \emph{schema rewrite} is a view on the database that removes tables before
fusion (\cref{sec:extraction}).
Every remaining table costs one forward pass of the tabular model.
The views read the temporally filtered tables, so RelICL applies the rewrites
once, and each timestamp only changes the filter.

\textbf{Inlining.}
RelICL left-joins leaf tables into every table that references it and then drops
it, which adds columns but no rows.
In an entity task in RelBench, every task row references one \emph{entity table}
that describes the entity the prediction is about.
RelICL inlines that table into the task table and rewires its remaining
references to the task table.

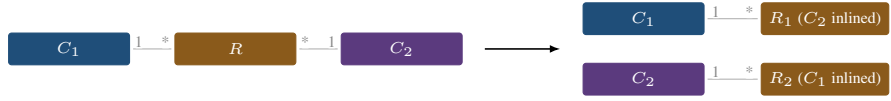
\begin{figure}
	\centering
	\begingroup
\def\cxA{0}       %
\def\cxB{22}      %
\def\cxC{44}      %
\def\cxD{76}      %
\def\cxE{100}     %
\def\cyU{4}       %
\def\cyL{-4}      %

\begin{tikzpicture}[
	x=1mm, y=1mm,
	rounded corners=0.4mm,
	tnode/.style={draw=#1, fill=#1, text=primarytext, font=\tiny,
			anchor=center, minimum width=16mm, minimum height=4.2mm,
			inner xsep=1mm},
	card/.style={font=\tiny, text=muted, inner sep=0.5pt, fill=white},
	fk/.style={border, shorten >=0.4mm, shorten <=0.4mm},
	flow/.style={-{Latex[length=1.4mm]}, semithick},
	]
	\node[tnode=secondary] (bc1) at (\cxA, 0) {$C_1$};
	\node[tnode=quaternary] (br) at (\cxB, 0) {$R$};
	\node[tnode=tertiary] (bc2) at (\cxC, 0) {$C_2$};
	\draw[fk] (br) -- node[card, above, pos=0.2] {*}
	node[card, above, pos=0.8] {1} (bc1);
	\draw[fk] (br) -- node[card, above, pos=0.2] {*}
	node[card, above, pos=0.8] {1} (bc2);

	\draw[flow] (\cxC+11, 0) -- (\cxD-11, 0);

	\node[tnode=secondary] (ac1) at (\cxD, \cyU) {$C_1$};
	\node[tnode=quaternary] (ar1) at (\cxE, \cyU) {$R_1$ ($C_2$ inlined)};
	\node[tnode=tertiary] (ac2) at (\cxD, \cyL) {$C_2$};
	\node[tnode=quaternary] (ar2) at (\cxE, \cyL) {$R_2$ ($C_1$ inlined)};
	\draw[fk] (ar1) -- node[card, above, pos=0.2] {*}
	node[card, above, pos=0.8] {1} (ac1);
	\draw[fk] (ar2) -- node[card, above, pos=0.2] {*}
	node[card, above, pos=0.8] {1} (ac2);
\end{tikzpicture}
\endgroup
	\caption{Junction table elimination.
		The junction table $R$ is on the many side of its references to $C_1$
		and $C_2$.
		RelICL replaces it by two copies, each of which keeps one reference and
		carries the columns of the other neighbor inline.
	}
	\label{fig:junction}
\end{figure}

\textbf{Junction tables.}
A \emph{junction table} $R$ references two tables $C_1$ and $C_2$ and is
referenced by none.
RelICL replaces $R$ by two copies (\cref{fig:junction}).
The copy $R_1$ carries the columns of $C_2$ inline and references $C_1$.
The copy $R_2$ carries the columns of $C_1$ inline and references $C_2$.
Pooling $R_1$ into $C_1$ then passes the attributes of $C_2$ to $C_1$ in one
step instead of two, and likewise for $R_2$.
This follows RelGNN \citep{Chen2025}, which also passes messages between
the two neighbors of a junction table in a single step.

\subsection{\Cref*{assm:embedding-sufficiency}: Row embeddings are good features}
\label{app:early-fusion}

This appendix supports \cref{assm:embedding-sufficiency}.

\textbf{Setup.}
We took ten binary classification datasets from OpenML \citep{Vanschoren2014},
listed in \Cref{tab:early-fusion-stats}.
Each one is a single table.
The setting $\tfm(R)$ ran TabICL on exactly that table $R$.
The alternative replaced all feature columns by a row embedding.
Both settings started from the same data, so $\tfm(R)$ is the upper bound this
construction can reach.
We ran the replacement twice.
The setting $\tfm(\tenc(R))$ appended the encoder output as it is, at
$d = 512$.
The setting $\tfm(\operatorname{reduce}(\tenc(R)))$ then applied the
Gaussian random projection that RelICL used in the main paper, at $d' = 32$
(see \cref{app:hpo}).
Every configuration used a single estimator and ran with three seeds.
Each seed determines both the stratified 80/20 split and the models' randomness.

\begin{table}
	\caption{%
		The ten OpenML datasets of \cref{app:early-fusion}.
	}
	\label{tab:early-fusion-stats}
	\centering
	\small
	\begin{tabular}{lrrrr}
\toprule
\textbf{Dataset} & \textbf{OpenML ID} & \textbf{Rows} & \textbf{Features} & \textbf{Positive class (\%)} \\
\midrule
bioresponse & 4134 & 3751 & 1776 & 54.2 \\
churn & 40701 & 5000 & 20 & 14.1 \\
compass & 42193 & 5278 & 13 & 47.0 \\
credit-g & 31 & 1000 & 20 & 70.0 \\
jasmine & 41143 & 2984 & 144 & 50.0 \\
kr-vs-kp & 3 & 3196 & 36 & 52.2 \\
philippine & 41145 & 5832 & 308 & 50.0 \\
phoneme & 1489 & 5404 & 5 & 29.3 \\
qsar-biodeg & 1494 & 1055 & 41 & 33.7 \\
spambase & 44 & 4601 & 57 & 39.4 \\
\bottomrule
\end{tabular}

\end{table}

\begin{table}
	\caption{%
		Replacing the feature columns by a row embedding (test AUROC (\%), higher
		is better, mean $\pm$ SEM over three seeds).
		$\tfm(R)$ runs TabICL on the table $R$.
		$\tfm(\tenc(R))$ replaces all feature columns by the row embedding
		($d = 512$).
		$\tfm(\operatorname{reduce}(\tenc(R)))$ then applies a Gaussian random
		projection ($d' = 32$, see
		\cref{app:hpo}).
		Both $\Delta$ are paired differences against the original data of the same
		seed.
		The last row averages the ten datasets, so its SEM is taken across
		datasets rather than across seeds.
	}
	\label{tab:early-fusion-full}
	\centering
	\footnotesize
	\begin{tabular}{lccccc}
\toprule
 & \textbf{$\tfm(R)$} & \multicolumn{2}{c}{\textbf{$\tfm(\tenc(R))$}} & \multicolumn{2}{c}{\textbf{$\tfm(\operatorname{reduce}(\tenc(R)))$}} \\
\cmidrule(lr){2-2}\cmidrule(lr){3-4}\cmidrule(lr){5-6}
\textbf{Dataset} & \textbf{AUROC} & \textbf{AUROC} & \textbf{$\Delta$} & \textbf{AUROC} & \textbf{$\Delta$} \\
\midrule
bioresponse & 87.7 $\pm$ 1.0 & 87.5 $\pm$ 0.8 & -0.3 $\pm$ 0.3 & 86.1 $\pm$ 1.0 & -1.6 $\pm$ 0.4 \\
churn & 92.8 $\pm$ 0.8 & 92.6 $\pm$ 0.6 & -0.2 $\pm$ 0.2 & 91.4 $\pm$ 0.6 & -1.4 $\pm$ 0.2 \\
compass & 72.3 $\pm$ 0.7 & 72.4 $\pm$ 0.7 & +0.1 $\pm$ 0.1 & 72.3 $\pm$ 0.7 & 0.0 $\pm$ 0.1 \\
credit-g & 82.2 $\pm$ 1.8 & 83.1 $\pm$ 0.4 & +0.9 $\pm$ 1.4 & 79.4 $\pm$ 0.4 & -2.8 $\pm$ 1.5 \\
jasmine & 89.8 $\pm$ 1.1 & 86.8 $\pm$ 0.4 & -3.0 $\pm$ 0.6 & 86.5 $\pm$ 0.8 & -3.3 $\pm$ 0.4 \\
kr-vs-kp & 100.0 $\pm$ 0.0 & 100.0 $\pm$ 0.0 & 0.0 $\pm$ 0.0 & 100.0 $\pm$ 0.0 & 0.0 $\pm$ 0.0 \\
philippine & 93.6 $\pm$ 0.1 & 84.7 $\pm$ 0.6 & -8.9 $\pm$ 0.6 & 83.5 $\pm$ 0.8 & -10.1 $\pm$ 0.8 \\
phoneme & 97.6 $\pm$ 0.0 & 96.2 $\pm$ 0.2 & -1.5 $\pm$ 0.2 & 96.2 $\pm$ 0.3 & -1.4 $\pm$ 0.3 \\
qsar-biodeg & 92.4 $\pm$ 2.1 & 91.3 $\pm$ 2.4 & -1.1 $\pm$ 0.4 & 91.1 $\pm$ 1.9 & -1.3 $\pm$ 0.3 \\
spambase & 99.1 $\pm$ 0.0 & 98.8 $\pm$ 0.0 & -0.4 $\pm$ 0.0 & 98.8 $\pm$ 0.0 & -0.3 $\pm$ 0.0 \\
\midrule
\textbf{$\emptyset$} & 90.8 $\pm$ 2.7 & 89.3 $\pm$ 2.6 & -1.4 $\pm$ 0.9 & 88.5 $\pm$ 2.8 & -2.2 $\pm$ 0.9 \\
\bottomrule
\end{tabular}

\end{table}

\textbf{Results.}
\Cref{tab:early-fusion-full} reports test AUROC and the regret of the
replacement.
On four of the ten datasets the replacement cost nothing measurable.
Eight of the ten lost at most 1.5 points.
The outliers are philippine at 8.9 points and jasmine at 3.0.
Width alone does not explain them, since bioresponse is the widest table of the
ten at 1776 columns and lost nothing measurable.

\section{Experimental setup}
\label{app:setup}

\subsection{Dataset and task statistics}
\label{app:datasets}

\Cref{tab:dataset-stats} summarizes the seven RelBench databases used in our
experiments \citep{Robinson2024, Gu2026a}.
\Cref{tab:tasks-classification,tab:tasks-regression} report, per task, the
number of rows in each split.
The last column reports, on the test split, the positive rate
(classification) or the target's mean and standard deviation (regression).
Both task tables are restricted to the 21 entity tasks.
RelBench ships further task types (e.g., autocompletion or recommendation) that
we did not evaluate.

\begin{table}
	\caption{RelBench databases}
	\label{tab:dataset-stats}
	\centering
	\small
	\begin{tabular}{lrrr}
\toprule
\textbf{Dataset} & \textbf{Tables} & \textbf{Rows} & \textbf{Columns} \\
\midrule
rel-amazon & 3 & 23,218,245 & 15 \\
rel-avito & 8 & 24,653,915 & 42 \\
rel-event & 5 & 44,822,992 & 131 \\
rel-f1 & 9 & 97,606 & 67 \\
rel-hm & 3 & 16,931,173 & 37 \\
rel-stack & 7 & 5,399,818 & 51 \\
rel-trial & 15 & 5,852,157 & 140 \\
\bottomrule
\end{tabular}

\end{table}

\begin{table}
	\caption{RelBench's (binary) classification tasks}
	\label{tab:tasks-classification}
	\centering
	\small
	\begin{tabular}{llrrrr}
\toprule
\textbf{Dataset} & \textbf{Task} & \textbf{Train} & \textbf{Val} & \textbf{Test} & \textbf{Pos. rate (\%)} \\
\midrule
rel-amazon & item-churn & 2,536,014 & 177,689 & 166,842 & 36.9 \\
rel-amazon & user-churn & 4,708,383 & 409,792 & 351,885 & 60.6 \\
rel-avito & user-clicks & 59,454 & 21,183 & 47,996 & 1.5 \\
rel-avito & user-visits & 86,619 & 29,979 & 36,129 & 85.1 \\
rel-event & user-ignore & 19,239 & 2,013 & 1,958 & 13.0 \\
rel-event & user-repeat & 3,842 & 268 & 246 & 44.7 \\
rel-f1 & driver-dnf & 11,411 & 566 & 702 & 70.5 \\
rel-f1 & driver-top3 & 1,353 & 588 & 726 & 17.6 \\
rel-hm & user-churn & 3,832,692 & 76,556 & 74,575 & 82.6 \\
rel-stack & user-badge & 3,386,276 & 247,398 & 255,360 & 2.6 \\
rel-stack & user-engagement & 1,360,850 & 85,838 & 88,137 & 2.7 \\
rel-trial & study-outcome & 11,994 & 960 & 825 & 58.5 \\
\bottomrule
\end{tabular}

\end{table}

\begin{table}
	\caption{RelBench's regression tasks}
	\label{tab:tasks-regression}
	\centering
	\small
	\begin{tabular}{llrrrr@{ $\pm$ }l}
\toprule
\textbf{Dataset} & \textbf{Task} & \textbf{Train} & \textbf{Val} & \textbf{Test} & \multicolumn{2}{c}{\textbf{Target (mean $\pm$ SD)}} \\
\midrule
rel-amazon & item-ltv & 2,707,679 & 166,978 & 178,334 & 77.13 & 664.61 \\
rel-amazon & user-ltv & 4,708,383 & 409,792 & 351,885 & 16.78 & 55.44 \\
rel-avito & ad-ctr & 5,100 & 1,766 & 1,816 & 0.05 & 0.11 \\
rel-event & user-attendance & 19,239 & 2,013 & 1,958 & 0.26 & 0.64 \\
rel-f1 & driver-position & 7,453 & 499 & 760 & 11.93 & 5.21 \\
rel-hm & item-sales & 5,488,184 & 105,542 & 105,542 & 0.08 & 0.58 \\
rel-stack & post-votes & 2,453,921 & 156,216 & 160,903 & 0.07 & 0.37 \\
rel-trial & site-success & 151,407 & 19,740 & 22,617 & 0.46 & 0.48 \\
rel-trial & study-adverse & 43,335 & 3,596 & 3,098 & 57.93 & 249.73 \\
\bottomrule
\end{tabular}

\end{table}

\subsection{Baselines}
\label{app:baselines}

The published numbers in \cref{sec:experiments,app:all-models} stem from
GraphSAGE \citep{Robinson2024}, RelGNN \citep{Chen2025}, RelGT
\citep{Dwivedi2025b}, RT \citep{Ranjan2025a}, PluRel \citep{Kothapalli2026},
KumoRFM-2 and DS+LightGBM \citep{Hudovernik2026a}, RDBLearn \citep{Xu2026a},
and TabPFN-Rel \citep{Hayler2026}.
RT (target DB) is RT after continued pretraining on the target database.
The papers differ in tuning, seeds, and evaluation protocol, so the tables give
only a rough comparison.

\subsection{Hyperparameter optimization}
\label{app:hpo}

We ran a simple hyperparameter optimization (HPO) scheme to determine a good
hyperparameter configuration for the tasks in RelBench.

\textbf{Tasks.}
The search ran independently on six classification and six regression tasks of
the 21 RelBench entity tasks (see \cref{tab:hpo-tasks}).
The other 9 of the 21 tasks were not used to find a suitable default setting.

\begin{table}
	\caption{Tasks the hyperparameter search ran on.}
	\label{tab:hpo-tasks}
	\centering
	\small
	\begin{tabular}{ll}
	\toprule
	\textbf{Regression}                & \textbf{Classification}          \\
	\midrule
	\texttt{rel-f1/driver-position}    & \texttt{rel-f1/driver-top3}      \\
	\texttt{rel-event/user-attendance} & \texttt{rel-event/user-repeat}   \\
	\texttt{rel-trial/study-adverse}   & \texttt{rel-event/user-ignore}   \\
	\texttt{rel-trial/site-success}    & \texttt{rel-trial/study-outcome} \\
	\texttt{rel-avito/ad-ctr}          & \texttt{rel-avito/user-clicks}   \\
	\texttt{rel-hm/item-sales}         & \texttt{rel-hm/user-churn}       \\
	\bottomrule
\end{tabular}

\end{table}

\textbf{Search space.}
\Cref{tab:hpo-space} lists all 11 hyperparameters, their search
space, and the value selected for the configuration RelICL reports
throughout the paper, unless declared otherwise.

\textbf{Selection.}
We chose each hyperparameter on its own rather than taking the best single
trial.
Within each task, we ranked the trials by validation score, since raw scores
are not comparable across tasks \citep{Demsar2006,Bardenet2013}.
For each value of a hyperparameter, we averaged these ranks over all trials
and tasks that used this value.
This averages over the other hyperparameters, similar to the marginals of
\citet{Hutter2014a}.
We then kept the value with the best mean rank (last column in
\cref{tab:hpo-space}).

\textbf{Regression specifics.}
The target transform was the one exception kept per task type: it is
undefined for classification, so it was fixed to \texttt{none} there.

\begin{table}
	\caption{Hyperparameters, their search space, and the value selected for
		RelICL's reported configuration.}
	\label{tab:hpo-space}
	\centering
	\small
	\begin{threeparttable}
		\begin{tabular}{lll}
	\toprule
	\textbf{Hyperparameter}                                        & \textbf{Search space}                     & \textbf{Selected (default values)} \\
	\midrule
	Anchor budget fraction\tnote{a}                                & $\{0.25, 0.75\}$                          & $0.75$                          \\
	Train sample size                                              & $\{16384, 32768, 65536\}$                 & $32768$                         \\
	Context table sample size                                      & $\{32768, 65536\}$                        & $65536$                         \\
	Count features after optimization\tnote{b}                     & $\{\text{true}, \text{false}\}$           & true                            \\
	Count feature windows (days)\tnote{b}                          &
	$\{\text{none}, [1,7,28], [1,3,7,28]\}$                        & none                                                                        \\
	Relative time encoding                                         & $\{\text{true}, \text{false}\}$           & true                            \\
	Early-fusion reduction dim.                                    & $\{32, 64\}$                              & $32$                            \\
	Early-fusion reduction method                                  &
	$\{\text{GRP}\tnote{c}, \text{PCA}, \text{first-}k\tnote{d}\}$ & GRP\tnote{c}                                                                \\
	Key pooling\tnote{e}                                           & $\{\text{mean}, \text{max}, \text{sum}\}$ & mean                            \\
	Outlier clipping threshold\tnote{f}                            & $\{4.0, 8.0\}$                            & $4.0$                           \\
	Target transform\tnote{*}                                      & $\{\text{none}, \text{log1p}\}$           & log1p / none                    \\
	\bottomrule
\end{tabular}

		\begin{tablenotes}
			\small
			\item[a] Share of the training-row budget reserved for rows
			sharing a key with the test row.
			\item[b] Adds per-key counts of referencing rows as extra
			features, optionally split into rolling windows over the last
			$N$ days.
			\item[c] GRP: Gaussian random projection
			\citep{Johnson1984,Dasgupta2003}.
			\item[d] first-$k$: keep the first $k$ embedding dimensions,
			dropping the rest.
			\item[e] How row embeddings sharing a key are aggregated into
			one.
			\item[f] TabICL specific.
			\item[*] Only defined for regression; fixed to \texttt{none}
			for classification.
		\end{tablenotes}
	\end{threeparttable}
\end{table}

\textbf{Run time.}
Across the 12 studies, the search consumed 121 GPU-hours in total (1200
completed trials), with a mean trial duration of 361 s (median 303 s).
Per-task mean trial duration ranged from 55 s (\texttt{rel-avito/ad-ctr})
to 887 s (\texttt{rel-hm/item-sales}).

\subsection{Configuration ensembling}
\label{app:ensembling}

The reported results in \cref{tab:regression-main,tab:classification-main}
ensembled over configurations.
Concretely, this combined per task type a small, fixed set of configuration
variants with the \emph{default values}, i.e., the configuration selected
by the hyperparameter search (\cref{app:hpo}).

\textbf{Configurations.}
Regression tasks combined the default values with one variant (2 members);
classification tasks combined the default values with three (4 members).
The same combination was used for every task of a given type: selection below is
joint across all tasks of that type, not selected individually per task.
Each variant changed exactly one setting from the default values.
For regression, the variant adds windowed count features over the last 1, 7,
and 28 days.
For classification, the first variant drops the reference count features
entirely.
The second disables the schema rewrites (\cref{app:schema-rewrites}).
The third reduces the embeddings of each backbone estimator separately instead
of merging all estimators before reducing.

\section{Additional results}
\label{app:results}

\subsection{All published models}
\label{app:all-models}

\Cref{tab:all-models-regression,tab:all-models-classification} compare RelICL
with every model for which we collected published RelBench results.

\begin{table}
	\caption{All published models, regression, MAE (lower is better) / $R^2$
		(\%, higher is better)}
	\label{tab:all-models-regression}
	\centering
	\resizebox{\linewidth}{!}{%
		\begin{threeparttable}
			\begin{tabular}{llr@{\;/\;}lr@{\;/\;}lr@{\;/\;}lr@{\;/\;}lr@{\;/\;}lr@{\;/\;}lr@{\;/\;}lr@{\;/\;}lr@{\;/\;}l}
\toprule
\textbf{Kind} & \textbf{Model} & \multicolumn{2}{c}{\rotatebox[origin=l]{90}{\textbf{item-ltv}}} & \multicolumn{2}{c}{\rotatebox[origin=l]{90}{\textbf{user-ltv}}} & \multicolumn{2}{c}{\rotatebox[origin=l]{90}{\textbf{ad-ctr}}} & \multicolumn{2}{c}{\rotatebox[origin=l]{90}{\textbf{user-attendance}}} & \multicolumn{2}{c}{\rotatebox[origin=l]{90}{\textbf{driver-position}}} & \multicolumn{2}{c}{\rotatebox[origin=l]{90}{\textbf{item-sales}}} & \multicolumn{2}{c}{\rotatebox[origin=l]{90}{\textbf{post-votes}}} & \multicolumn{2}{c}{\rotatebox[origin=l]{90}{\textbf{site-success}}} & \multicolumn{2}{c}{\rotatebox[origin=l]{90}{\textbf{study-adverse}}} \\
\midrule
Trained & GraphSAGE & 50.05 & -- & 14.31 & -- & 0.04 & -- & 0.26 & -- & 4.02 & -- & 0.06 & -- & 0.07 & -- & 0.40 & -- & 44.47 & -- \\
Trained & RelGNN & 48.77 & -- & 14.23 & -- & 0.04 & -- & 0.24 & -- & 3.80 & -- & 0.05 & -- & 0.07 & -- & \textbf{0.30} & -- & 44.46 & -- \\
Trained & RelGT & 48.92 & -- & 14.27 & -- & 0.04 & -- & 0.25 & -- & 3.92 & -- & 0.05 & -- & 0.07 & -- & 0.33 & -- & 43.99 & -- \\
Trained & DS+LightGBM & \textbf{41.12} & -- & 13.93 & -- & 0.04 & -- & 0.28 & -- & 3.96 & -- & 0.04 & -- & 0.07 & -- & 0.41 & -- & 40.58 & -- \\
Pretrained (rel.) & RT & -- & 32.5 & -- & 36.9 & -- & 4.5 & \multicolumn{2}{c}{--} & -- & 52.4 & -- & 14.0 & -- & 33.9 & -- & 4.5 & -- & 2.6 \\
Pretrained (rel.) & RT (target DB) & -- & 32.2 & -- & \textbf{38.3} & -- & 8.0 & \multicolumn{2}{c}{--} & -- & \textbf{58.7} & -- & 30.9 & -- & \textbf{35.0} & -- & 5.1 & -- & 3.1 \\
Pretrained (rel.) & PluRel & -- & \textbf{40.5} & -- & 18.5 & -- & 4.9 & \multicolumn{2}{c}{--} & -- & 55.5 & -- & 20.0 & -- & 25.5 & -- & \textbf{38.6} & -- & 1.6 \\
Pretrained (rel.) & KumoRFM-2 & 46.99 & -- & \textbf{13.92} & -- & \textbf{0.03} & -- & \textbf{0.23} & -- & \textbf{2.85} & -- & \textbf{0.03} & -- & \textbf{0.06} & -- & 0.43 & -- & 43.29 & -- \\
Pretrained (tab.) & RDBLearn & 48.56 & -- & 14.54 & -- & \textbf{0.03} & -- & \multicolumn{2}{c}{--} & \multicolumn{2}{c}{--} & 0.06 & -- & 0.07 & -- & 0.42 & -- & 43.91 & -- \\
Pretrained (tab.) & TabPFN-Rel (API) & 46.80 & -- & 14.40 & -- & \textbf{0.03} & -- & 0.24 & -- & 3.77 & -- & 0.06 & -- & 0.07 & -- & 0.41 & -- & 39.80 & -- \\
Pretrained (tab.) & TabPFN-Rel & 47.80 & -- & 14.40 & -- & \textbf{0.03} & -- & 0.24 & -- & 3.76 & -- & 0.06 & -- & 0.07 & -- & 0.39 & -- & 42.60 & -- \\
Pretrained (tab.) & RelICL (TabICL) & 43.97 & 28.6 & 14.78 & 21.0 & 0.04 & \textbf{18.3} & 0.24 & 5.1 & 3.68 & 25.2 & 0.04 & \textbf{69.2} & \textbf{0.06} & 25.6 & 0.40 & -22.0 & 32.42 & 54.0 \\
Pretrained (tab.) & RelICL (TabPFN) & 43.41 & 29.2 & 15.48 & 16.5 & 0.05 & 0.1 & 0.30 & \textbf{17.2} & 4.00 & 13.4 & 0.04 & 50.1 & 0.08 & 17.0 & 0.43 & -7.1 & \textbf{29.97} & \textbf{56.9} \\
\bottomrule
\end{tabular}

			\begin{tablenotes}
				\item[] -- Not reported.
			\end{tablenotes}
		\end{threeparttable}}
\end{table}

\begin{table}
	\caption{All published models, classification (AUROC (\%), higher is
		better)}
	\label{tab:all-models-classification}
	\centering
	\resizebox{\linewidth}{!}{%
		\begin{threeparttable}
			\begin{tabular}{llcccccccccccc}
\toprule
\textbf{Kind} & \textbf{Model} & \rotatebox[origin=l]{90}{\textbf{item-churn}} & \rotatebox[origin=l]{90}{\textbf{user-churn (ama)}} & \rotatebox[origin=l]{90}{\textbf{user-clicks}} & \rotatebox[origin=l]{90}{\textbf{user-visits}} & \rotatebox[origin=l]{90}{\textbf{user-ignore}} & \rotatebox[origin=l]{90}{\textbf{user-repeat}} & \rotatebox[origin=l]{90}{\textbf{driver-dnf}} & \rotatebox[origin=l]{90}{\textbf{driver-top3}} & \rotatebox[origin=l]{90}{\textbf{user-churn (hm)}} & \rotatebox[origin=l]{90}{\textbf{user-badge}} & \rotatebox[origin=l]{90}{\textbf{user-engagement}} & \rotatebox[origin=l]{90}{\textbf{study-outcome}} \\
\midrule
Trained & GraphSAGE & \textbf{82.8} & 70.4 & 65.9 & 66.2 & 81.6 & 76.9 & 72.6 & 75.5 & 69.9 & 88.9 & 90.6 & 68.6 \\
Trained & RelGNN & 82.6 & \textbf{71.0} & 68.2 & 66.2 & 86.2 & 79.6 & 75.3 & 85.7 & \textbf{70.9} & \textbf{89.0} & \textbf{90.8} & 71.2 \\
Trained & RelGT & 82.5 & 70.4 & 68.3 & 66.8 & 81.6 & 76.1 & 75.9 & 83.5 & 69.3 & 86.3 & 90.5 & 68.6 \\
Trained & DS+LightGBM & 81.8 & 67.6 & 64.3 & 64.5 & 84.2 & 75.4 & 69.8 & 82.4 & 69.0 & 86.2 & 90.3 & 72.0 \\
Pretrained (rel.) & RT & 70.9 & 64.0 & 59.5 & 61.8 & -- & -- & 81.2 & 89.3 & 62.8 & 80.1 & 75.7 & 51.8 \\
Pretrained (rel.) & PluRel & 72.5 & 65.0 & 47.9 & 63.4 & -- & -- & 81.0 & 88.4 & 66.0 & 82.0 & 86.2 & 51.8 \\
Pretrained (rel.) & KumoRFM-2 & 82.2 & 69.1 & 67.4 & \textbf{69.4} & \textbf{90.8} & \textbf{81.7} & \textbf{84.6} & \textbf{92.2} & 69.3 & 87.2 & 89.4 & 72.0 \\
Pretrained (tab.) & RDBLearn & 82.1 & 67.6 & \textbf{69.0} & 65.5 & -- & -- & -- & -- & 68.0 & 85.3 & 89.4 & 71.6 \\
Pretrained (tab.) & TabPFN-Rel (API) & \textbf{82.8} & 70.9 & 67.5 & 66.8 & 87.9 & 75.9 & 73.2 & 77.1 & 70.5 & 88.0 & 90.6 & \textbf{76.5} \\
Pretrained (tab.) & TabPFN-Rel & \textbf{82.8} & 70.2 & 61.5 & 66.9 & 70.1 & 76.9 & 71.5 & 79.3 & 70.6 & 86.4 & 90.6 & 73.1 \\
Pretrained (tab.) & RelICL (TabICL) & 78.5 & 65.4 & 64.3 & 66.2 & 88.5 & 79.3 & 73.6 & 81.4 & 68.4 & 86.3 & 90.0 & 73.4 \\
Pretrained (tab.) & RelICL (TabPFN) & 78.4 & 64.6 & 58.7 & 65.7 & 87.9 & 77.7 & 74.2 & 78.4 & 67.8 & 84.9 & 89.5 & 73.6 \\
\bottomrule
\end{tabular}

			\begin{tablenotes}
				\item[] -- Not reported.
			\end{tablenotes}
		\end{threeparttable}}
\end{table}

\subsection{Head-to-head wins}
\label{app:head-to-head}

\Cref{tab:win-matrix-regression,tab:win-matrix-classification} count wins pair
by pair over the models of \cref{tab:regression-main,tab:classification-main},
and \cref{tab:win-matrix-all-regression,tab:win-matrix-all-classification} do
the same over every published model.
Regression compares on MAE where both models report it and on $R^2$ otherwise,
because RT and PluRel report only $R^2$ while the trained baselines report only
MAE.
An asterisk marks a cell in which $R^2$ decided at least one task.
A tie counts as a win for neither model, so the two cells of a pair can sum to
less than the number of shared tasks.
Bold marks a pair the row model wins; ties are underlined.

\begin{table}
	\caption{Head-to-head wins, regression}
	\label{tab:win-matrix-regression}
	\centering
	\small
	\begin{threeparttable}
		\begin{tabular}{llccccc}
\toprule
\multicolumn{2}{c}{} & \multicolumn{2}{c}{\textbf{Pretrained (rel.)}} & \multicolumn{3}{c}{\textbf{Pretrained (tab.)}} \\
\cmidrule(lr){3-4}\cmidrule(lr){5-7}
\textbf{Kind} & \textbf{Model} & \rotatebox[origin=l]{90}{\textbf{RT}} & \rotatebox[origin=l]{90}{\textbf{PluRel}} & \rotatebox[origin=l]{90}{\textbf{RDBLearn}} & \rotatebox[origin=l]{90}{\textbf{TabPFN-Rel}} & \rotatebox[origin=l]{90}{\textbf{RelICL}} \\
\midrule
Pretrained (rel.) & RT & -- & 3\,/\,8\rlap{$^{*}$} & -- & -- & \textbf{5\,/\,8}\rlap{$^{*}$} \\
Pretrained (rel.) & PluRel & \textbf{5\,/\,8}\rlap{$^{*}$} & -- & -- & -- & 3\,/\,8\rlap{$^{*}$} \\
Pretrained (tab.) & RDBLearn & -- & -- & -- & 0\,/\,7 & 2\,/\,7 \\
Pretrained (tab.) & TabPFN-Rel & -- & -- & \textbf{6\,/\,7} & -- & 4\,/\,9 \\
Pretrained (tab.) & RelICL & 3\,/\,8\rlap{$^{*}$} & \textbf{5\,/\,8}\rlap{$^{*}$} & \textbf{5\,/\,7} & \textbf{5\,/\,9} & -- \\
\bottomrule
\end{tabular}

		\begin{tablenotes}
			\item[*] $R^2$ decided at least one task of this cell.
			\item[--] Same model, or no shared task.
		\end{tablenotes}
	\end{threeparttable}
\end{table}

\begin{table}
	\caption{Head-to-head wins, classification}
	\label{tab:win-matrix-classification}
	\centering
	\small
	\begin{threeparttable}
		\begin{tabular}{llccccc}
\toprule
\multicolumn{2}{c}{} & \multicolumn{2}{c}{\textbf{Pretrained (rel.)}} & \multicolumn{3}{c}{\textbf{Pretrained (tab.)}} \\
\cmidrule(lr){3-4}\cmidrule(lr){5-7}
\textbf{Kind} & \textbf{Model} & \rotatebox[origin=l]{90}{\textbf{RT}} & \rotatebox[origin=l]{90}{\textbf{PluRel}} & \rotatebox[origin=l]{90}{\textbf{RDBLearn}} & \rotatebox[origin=l]{90}{\textbf{TabPFN-Rel}} & \rotatebox[origin=l]{90}{\textbf{RelICL}} \\
\midrule
Pretrained (rel.) & RT & -- & \phantom{0}3\,/\,10 & \phantom{0}0\,/\,8\phantom{0} & \phantom{0}2\,/\,10 & \phantom{0}2\,/\,10 \\
Pretrained (rel.) & PluRel & \textbf{\phantom{0}6\,/\,10} & -- & \phantom{0}0\,/\,8\phantom{0} & \phantom{0}2\,/\,10 & \phantom{0}2\,/\,10 \\
Pretrained (tab.) & RDBLearn & \textbf{\phantom{0}8\,/\,8\phantom{0}} & \textbf{\phantom{0}8\,/\,8\phantom{0}} & -- & \phantom{0}1\,/\,8\phantom{0} & \phantom{0}3\,/\,8\phantom{0} \\
Pretrained (tab.) & TabPFN-Rel & \textbf{\phantom{0}8\,/\,10} & \textbf{\phantom{0}8\,/\,10} & \textbf{\phantom{0}7\,/\,8\phantom{0}} & -- & \underline{\phantom{0}6\,/\,12} \\
Pretrained (tab.) & RelICL & \textbf{\phantom{0}8\,/\,10} & \textbf{\phantom{0}8\,/\,10} & \textbf{\phantom{0}5\,/\,8\phantom{0}} & \underline{\phantom{0}6\,/\,12} & -- \\
\bottomrule
\end{tabular}

		\begin{tablenotes}
			\item[] -- Same model.
		\end{tablenotes}
	\end{threeparttable}
\end{table}

\begin{table}
	\caption{Head-to-head wins against all published models, regression}
	\label{tab:win-matrix-all-regression}
	\centering
	\resizebox{\linewidth}{!}{%
		\begin{threeparttable}
			\begin{tabular}{llcccccccccccc}
\toprule
\multicolumn{2}{c}{} & \multicolumn{4}{c}{\textbf{Trained}} & \multicolumn{4}{c}{\textbf{Pretrained (rel.)}} & \multicolumn{4}{c}{\textbf{Pretrained (tab.)}} \\
\cmidrule(lr){3-6}\cmidrule(lr){7-10}\cmidrule(lr){11-14}
\textbf{Kind} & \textbf{Model} & \rotatebox[origin=l]{90}{\textbf{GraphSAGE}} & \rotatebox[origin=l]{90}{\textbf{RelGNN}} & \rotatebox[origin=l]{90}{\textbf{RelGT}} & \rotatebox[origin=l]{90}{\textbf{DS+LightGBM}} & \rotatebox[origin=l]{90}{\textbf{RT}} & \rotatebox[origin=l]{90}{\textbf{RT (target DB)}} & \rotatebox[origin=l]{90}{\textbf{PluRel}} & \rotatebox[origin=l]{90}{\textbf{KumoRFM-2}} & \rotatebox[origin=l]{90}{\textbf{RDBLearn}} & \rotatebox[origin=l]{90}{\textbf{TabPFN-Rel (API)}} & \rotatebox[origin=l]{90}{\textbf{TabPFN-Rel}} & \rotatebox[origin=l]{90}{\textbf{RelICL}} \\
\midrule
Trained & GraphSAGE & -- & \phantom{0}0\,/\,9\phantom{0} & \phantom{0}0\,/\,9\phantom{0} & \phantom{0}3\,/\,9\phantom{0} & -- & -- & -- & \phantom{0}1\,/\,9\phantom{0} & \textbf{\phantom{0}4\,/\,7\phantom{0}} & \phantom{0}4\,/\,9\phantom{0} & \phantom{0}3\,/\,9\phantom{0} & \phantom{0}1\,/\,9\phantom{0} \\
Trained & RelGNN & \textbf{\phantom{0}8\,/\,9\phantom{0}} & -- & \textbf{\phantom{0}5\,/\,9\phantom{0}} & \underline{\phantom{0}4\,/\,9\phantom{0}} & -- & -- & -- & \phantom{0}1\,/\,9\phantom{0} & \textbf{\phantom{0}4\,/\,7\phantom{0}} & \textbf{\phantom{0}5\,/\,9\phantom{0}} & \textbf{\phantom{0}5\,/\,9\phantom{0}} & \phantom{0}4\,/\,9\phantom{0} \\
Trained & RelGT & \textbf{\phantom{0}8\,/\,9\phantom{0}} & \phantom{0}2\,/\,9\phantom{0} & -- & \underline{\phantom{0}4\,/\,9\phantom{0}} & -- & -- & -- & \phantom{0}1\,/\,9\phantom{0} & \textbf{\phantom{0}4\,/\,7\phantom{0}} & \phantom{0}4\,/\,9\phantom{0} & \phantom{0}4\,/\,9\phantom{0} & \phantom{0}3\,/\,9\phantom{0} \\
Trained & DS+LightGBM & \textbf{\phantom{0}5\,/\,9\phantom{0}} & \underline{\phantom{0}4\,/\,9\phantom{0}} & \underline{\phantom{0}4\,/\,9\phantom{0}} & -- & -- & -- & -- & \phantom{0}3\,/\,9\phantom{0} & \textbf{\phantom{0}6\,/\,7\phantom{0}} & \textbf{\phantom{0}5\,/\,9\phantom{0}} & \textbf{\phantom{0}5\,/\,9\phantom{0}} & \phantom{0}3\,/\,9\phantom{0} \\
Pretrained (rel.) & RT & -- & -- & -- & -- & -- & \phantom{0}1\,/\,8\phantom{0}\rlap{$^{*}$} & \phantom{0}3\,/\,8\phantom{0}\rlap{$^{*}$} & -- & -- & -- & -- & \textbf{\phantom{0}5\,/\,8\phantom{0}}\rlap{$^{*}$} \\
Pretrained (rel.) & RT (target DB) & -- & -- & -- & -- & \textbf{\phantom{0}7\,/\,8\phantom{0}}\rlap{$^{*}$} & -- & \textbf{\phantom{0}6\,/\,8\phantom{0}}\rlap{$^{*}$} & -- & -- & -- & -- & \textbf{\phantom{0}5\,/\,8\phantom{0}}\rlap{$^{*}$} \\
Pretrained (rel.) & PluRel & -- & -- & -- & -- & \textbf{\phantom{0}5\,/\,8\phantom{0}}\rlap{$^{*}$} & \phantom{0}2\,/\,8\phantom{0}\rlap{$^{*}$} & -- & -- & -- & -- & -- & \phantom{0}3\,/\,8\phantom{0}\rlap{$^{*}$} \\
Pretrained (rel.) & KumoRFM-2 & \textbf{\phantom{0}8\,/\,9\phantom{0}} & \textbf{\phantom{0}8\,/\,9\phantom{0}} & \textbf{\phantom{0}8\,/\,9\phantom{0}} & \textbf{\phantom{0}6\,/\,9\phantom{0}} & -- & -- & -- & -- & \textbf{\phantom{0}5\,/\,7\phantom{0}} & \textbf{\phantom{0}5\,/\,9\phantom{0}} & \textbf{\phantom{0}6\,/\,9\phantom{0}} & \textbf{\phantom{0}6\,/\,9\phantom{0}} \\
Pretrained (tab.) & RDBLearn & \phantom{0}3\,/\,7\phantom{0} & \phantom{0}3\,/\,7\phantom{0} & \phantom{0}3\,/\,7\phantom{0} & \phantom{0}1\,/\,7\phantom{0} & -- & -- & -- & \phantom{0}1\,/\,7\phantom{0} & -- & \phantom{0}0\,/\,7\phantom{0} & \phantom{0}0\,/\,7\phantom{0} & \phantom{0}2\,/\,7\phantom{0} \\
Pretrained (tab.) & TabPFN-Rel (API) & \textbf{\phantom{0}5\,/\,9\phantom{0}} & \phantom{0}4\,/\,9\phantom{0} & \textbf{\phantom{0}5\,/\,9\phantom{0}} & \phantom{0}4\,/\,9\phantom{0} & -- & -- & -- & \phantom{0}4\,/\,9\phantom{0} & \textbf{\phantom{0}7\,/\,7\phantom{0}} & -- & \textbf{\phantom{0}5\,/\,9\phantom{0}} & \phantom{0}3\,/\,9\phantom{0} \\
Pretrained (tab.) & TabPFN-Rel & \textbf{\phantom{0}6\,/\,9\phantom{0}} & \phantom{0}4\,/\,9\phantom{0} & \textbf{\phantom{0}5\,/\,9\phantom{0}} & \phantom{0}4\,/\,9\phantom{0} & -- & -- & -- & \phantom{0}3\,/\,9\phantom{0} & \textbf{\phantom{0}6\,/\,7\phantom{0}} & \phantom{0}3\,/\,9\phantom{0} & -- & \phantom{0}4\,/\,9\phantom{0} \\
Pretrained (tab.) & RelICL & \textbf{\phantom{0}8\,/\,9\phantom{0}} & \textbf{\phantom{0}5\,/\,9\phantom{0}} & \textbf{\phantom{0}6\,/\,9\phantom{0}} & \textbf{\phantom{0}6\,/\,9\phantom{0}} & \phantom{0}3\,/\,8\phantom{0}\rlap{$^{*}$} & \phantom{0}3\,/\,8\phantom{0}\rlap{$^{*}$} & \textbf{\phantom{0}5\,/\,8\phantom{0}}\rlap{$^{*}$} & \phantom{0}3\,/\,9\phantom{0} & \textbf{\phantom{0}5\,/\,7\phantom{0}} & \textbf{\phantom{0}6\,/\,9\phantom{0}} & \textbf{\phantom{0}5\,/\,9\phantom{0}} & -- \\
\bottomrule
\end{tabular}

			\begin{tablenotes}
				\item[*] $R^2$ decided at least one task of this cell.
				\item[--] Same model, or no shared task.
			\end{tablenotes}
		\end{threeparttable}}
\end{table}

\begin{table}
	\caption{Head-to-head wins against all published models, classification}
	\label{tab:win-matrix-all-classification}
	\centering
	\resizebox{\linewidth}{!}{%
		\begin{threeparttable}
			\begin{tabular}{llccccccccccc}
\toprule
\multicolumn{2}{c}{} & \multicolumn{4}{c}{\textbf{Trained}} & \multicolumn{3}{c}{\textbf{Pretrained (rel.)}} & \multicolumn{4}{c}{\textbf{Pretrained (tab.)}} \\
\cmidrule(lr){3-6}\cmidrule(lr){7-9}\cmidrule(lr){10-13}
\textbf{Kind} & \textbf{Model} & \rotatebox[origin=l]{90}{\textbf{GraphSAGE}} & \rotatebox[origin=l]{90}{\textbf{RelGNN}} & \rotatebox[origin=l]{90}{\textbf{RelGT}} & \rotatebox[origin=l]{90}{\textbf{DS+LightGBM}} & \rotatebox[origin=l]{90}{\textbf{RT}} & \rotatebox[origin=l]{90}{\textbf{PluRel}} & \rotatebox[origin=l]{90}{\textbf{KumoRFM-2}} & \rotatebox[origin=l]{90}{\textbf{RDBLearn}} & \rotatebox[origin=l]{90}{\textbf{TabPFN-Rel (API)}} & \rotatebox[origin=l]{90}{\textbf{TabPFN-Rel}} & \rotatebox[origin=l]{90}{\textbf{RelICL}} \\
\midrule
Trained & GraphSAGE & -- & \phantom{0}2\,/\,12 & \textbf{\phantom{0}7\,/\,12} & \textbf{\phantom{0}9\,/\,12} & \textbf{\phantom{0}8\,/\,10} & \textbf{\phantom{0}8\,/\,10} & \phantom{0}5\,/\,12 & \textbf{\phantom{0}6\,/\,8\phantom{0}} & \phantom{0}3\,/\,12 & \textbf{\phantom{0}7\,/\,12} & \underline{\phantom{0}6\,/\,12} \\
Trained & RelGNN & \textbf{10\,/\,12} & -- & \textbf{\phantom{0}9\,/\,12} & \textbf{11\,/\,12} & \textbf{\phantom{0}8\,/\,10} & \textbf{\phantom{0}8\,/\,10} & \underline{\phantom{0}6\,/\,12} & \textbf{\phantom{0}6\,/\,8\phantom{0}} & \textbf{\phantom{0}8\,/\,12} & \textbf{\phantom{0}9\,/\,12} & \textbf{\phantom{0}9\,/\,12} \\
Trained & RelGT & \phantom{0}5\,/\,12 & \phantom{0}3\,/\,12 & -- & \textbf{10\,/\,12} & \textbf{\phantom{0}8\,/\,10} & \textbf{\phantom{0}8\,/\,10} & \phantom{0}4\,/\,12 & \textbf{\phantom{0}6\,/\,8\phantom{0}} & \phantom{0}4\,/\,12 & \phantom{0}5\,/\,12 & \textbf{\phantom{0}8\,/\,12} \\
Trained & DS+LightGBM & \phantom{0}3\,/\,12 & \phantom{0}1\,/\,12 & \phantom{0}2\,/\,12 & -- & \textbf{\phantom{0}8\,/\,10} & \textbf{\phantom{0}8\,/\,10} & \phantom{0}1\,/\,12 & \textbf{\phantom{0}5\,/\,8\phantom{0}} & \phantom{0}1\,/\,12 & \phantom{0}3\,/\,12 & \phantom{0}5\,/\,12 \\
Pretrained (rel.) & RT & \phantom{0}2\,/\,10 & \phantom{0}2\,/\,10 & \phantom{0}2\,/\,10 & \phantom{0}2\,/\,10 & -- & \phantom{0}3\,/\,10 & \phantom{0}0\,/\,10 & \phantom{0}0\,/\,8\phantom{0} & \phantom{0}2\,/\,10 & \phantom{0}2\,/\,10 & \phantom{0}2\,/\,10 \\
Pretrained (rel.) & PluRel & \phantom{0}2\,/\,10 & \phantom{0}2\,/\,10 & \phantom{0}2\,/\,10 & \phantom{0}2\,/\,10 & \textbf{\phantom{0}6\,/\,10} & -- & \phantom{0}0\,/\,10 & \phantom{0}0\,/\,8\phantom{0} & \phantom{0}2\,/\,10 & \phantom{0}2\,/\,10 & \phantom{0}2\,/\,10 \\
Pretrained (rel.) & KumoRFM-2 & \textbf{\phantom{0}7\,/\,12} & \underline{\phantom{0}6\,/\,12} & \textbf{\phantom{0}7\,/\,12} & \textbf{11\,/\,12} & \textbf{10\,/\,10} & \textbf{10\,/\,10} & -- & \textbf{\phantom{0}7\,/\,8\phantom{0}} & \phantom{0}5\,/\,12 & \textbf{\phantom{0}7\,/\,12} & \textbf{10\,/\,12} \\
Pretrained (tab.) & RDBLearn & \phantom{0}2\,/\,8\phantom{0} & \phantom{0}2\,/\,8\phantom{0} & \phantom{0}2\,/\,8\phantom{0} & \phantom{0}3\,/\,8\phantom{0} & \textbf{\phantom{0}8\,/\,8\phantom{0}} & \textbf{\phantom{0}8\,/\,8\phantom{0}} & \phantom{0}1\,/\,8\phantom{0} & -- & \phantom{0}1\,/\,8\phantom{0} & \phantom{0}1\,/\,8\phantom{0} & \phantom{0}3\,/\,8\phantom{0} \\
Pretrained (tab.) & TabPFN-Rel (API) & \textbf{\phantom{0}9\,/\,12} & \phantom{0}4\,/\,12 & \textbf{\phantom{0}8\,/\,12} & \textbf{11\,/\,12} & \textbf{\phantom{0}8\,/\,10} & \textbf{\phantom{0}8\,/\,10} & \textbf{\phantom{0}7\,/\,12} & \textbf{\phantom{0}7\,/\,8\phantom{0}} & -- & \textbf{\phantom{0}8\,/\,12} & \textbf{\phantom{0}8\,/\,12} \\
Pretrained (tab.) & TabPFN-Rel & \phantom{0}5\,/\,12 & \phantom{0}3\,/\,12 & \textbf{\phantom{0}7\,/\,12} & \textbf{\phantom{0}9\,/\,12} & \textbf{\phantom{0}8\,/\,10} & \textbf{\phantom{0}8\,/\,10} & \phantom{0}5\,/\,12 & \textbf{\phantom{0}7\,/\,8\phantom{0}} & \phantom{0}4\,/\,12 & -- & \underline{\phantom{0}6\,/\,12} \\
Pretrained (tab.) & RelICL & \underline{\phantom{0}6\,/\,12} & \phantom{0}3\,/\,12 & \phantom{0}4\,/\,12 & \textbf{\phantom{0}7\,/\,12} & \textbf{\phantom{0}8\,/\,10} & \textbf{\phantom{0}8\,/\,10} & \phantom{0}2\,/\,12 & \textbf{\phantom{0}5\,/\,8\phantom{0}} & \phantom{0}4\,/\,12 & \underline{\phantom{0}6\,/\,12} & -- \\
\bottomrule
\end{tabular}

			\begin{tablenotes}
				\item[] -- Same model.
			\end{tablenotes}
		\end{threeparttable}}
\end{table}

\end{document}